\documentclass[10pt,twocolumn,letterpaper, ]{article}

\usepackage{cvpr}              

\usepackage{graphicx}
\usepackage{amsmath}
\usepackage{amssymb}
\usepackage{booktabs}
\usepackage{multirow}
\usepackage{overpic}
\usepackage{amsmath}
\usepackage{wrapfig}
\usepackage{overpic}
\usepackage{bm}
\usepackage{multicol}
\usepackage{wrapfig}
\usepackage{algorithm}
\usepackage{algorithmic}
\usepackage{booktabs}       
\usepackage{amsfonts}       
\usepackage{nicefrac}       
\usepackage{microtype}      
\usepackage{color}
\newtheorem{theorem}{Theorem}
\newtheorem{definition}{Definition}

\usepackage[dvipsnames]{xcolor}
\definecolor{hollywoodcerise}{rgb}{0.96, 0.0, 0.63}
\definecolor{lasallegreen}{rgb}{0.03, 0.47, 0.19}
\definecolor{hanpurple}{rgb}{0.32, 0.09, 0.98}
\definecolor{green(pigment)}{rgb}{0.0, 0.65, 0.31}
\definecolor{yellow}{rgb}{0.85, 0.85, 0.31}
\definecolor{blue_light}{rgb}{0.34, 0.95, 0.95}
\usepackage[pagebackref=true,breaklinks=true,letterpaper=true,colorlinks,bookmarks=false]{hyperref}
\hypersetup{colorlinks,linkcolor={red},citecolor={hollywoodcerise},urlcolor={red}}  
\newcommand*{\affmark}[1][*]{\textsuperscript{#1}}

\usepackage[capitalize]{cleveref}
\crefname{section}{Sec.}{Secs.}
\Crefname{section}{Section}{Sections}
\Crefname{table}{Table}{Tables}
\crefname{table}{Tab.}{Tabs.}

\def\cvprPaperID{12593} 
\def\confName{CVPR}
\def\confYear{2023}
\usepackage[accsupp]{axessibility}
\begin{document}
\title{Patch-Mix Transformer for Unsupervised Domain Adaptation: A Game Perspective}

\author{Jinjing Zhu$^{1}$\thanks{These authors contributed equally to this work. }
\quad
Haotian Bai$^{1}$\footnotemark[1]
\quad
Lin Wang$^{1,2}$\thanks{Corresponding Author.}
\and
\affmark[1] AI Thrust, HKUST(GZ)\quad
\affmark[2] Dept. of CSE, HKUST\\
\quad
{\tt\small zhujinjing.hkust@gmail.com, haotianwhite@outlook.com, linwang@ust.hk}
}

\maketitle
\begin{abstract}
\vspace{-10pt}
Endeavors have been recently made to leverage the vision transformer (ViT) for the challenging unsupervised domain adaptation (UDA) task. They typically adopt the cross-attention in ViT for direct domain alignment. 
However, as the performance of cross-attention highly relies on the quality of pseudo labels for targeted samples, it becomes less effective when the domain gap becomes large. 
We solve this problem from a game theory's perspective with the proposed model dubbed as \textbf{PMTrans}, which bridges source and target domains with an intermediate domain. 
Specifically, we propose a novel ViT-based module called \textbf{PatchMix} that effectively builds up the intermediate domain,~\ie, probability distribution, by learning to sample patches from both domains based on the game-theoretical models.
This way, it learns to mix the patches from the source and target domains to maximize the cross entropy (CE), while exploiting two semi-supervised mixup losses in the feature and label spaces to minimize it.
As such, we interpret the process of UDA as a min-max CE game with three players, including the feature extractor, classifier, and PatchMix, to find the Nash Equilibria. 
Moreover, we leverage attention maps from ViT to re-weight the label of each patch by its importance, making it possible to obtain more domain-discriminative feature representations. We conduct extensive experiments on \textbf{four} benchmark datasets, and the results show that PMTrans significantly surpasses the ViT-based and CNN-based SoTA methods by \textbf{+3.6\%} on Office-Home, \textbf{+1.4\%} on Office-31, and \textbf{+7.2\%} on DomainNet, respectively. \url{https://vlis2022.github.io/cvpr23/PMTrans}
\end{abstract}
\vspace{-18pt}
\section{Introduction}
\vspace{-5pt}
\begin{figure}[t]
    \centering
    \captionsetup{font=small}
    \includegraphics[width=0.85\linewidth]{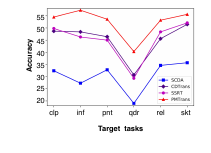}
    \vspace{-8pt}
    \caption{The classification accuracy of our PMTrans surpasses the SoTA methods by \textbf{+7.2\%} on the most challenging DomainNet dataset. Note that the target tasks treat one domain of DomainNet as the target domain and the others as the source domains.}
    \label{fig:result_cover}
    \vspace{-20pt}
\end{figure}
Convolutional neural networks (CNNs) have achieved tremendous success on numerous vision tasks; however, they still suffer from the limited generalization capability to a new domain due to the domain shift problem~\cite{ZhangDTZJ22}. Unsupervised domain adaptation (UDA) tackles this issue by transferring knowledge from a labeled source domain to an unlabeled target domain~\cite{PanY10}. A significant line of solutions reduces the domain gap based on the category-level alignment which produces pseudo labels for the target samples, such as metric learning~\cite{Kang0YH19, ZhuZWKCBXH21}, adversarial training~\cite{SaitoWUH18, Du0S0021, LiL0Y21}, and optimal transport~\cite{XuLWC020}. Furthermore, several works ~\cite{Dosovitskiy2021AnII,abs-2204-07683} explore the potential of ViT for the non-trivial UDA task. Recently, CDTrans \cite{abs-2109-06165} exploits the cross-attention in ViT for direct domain alignment, buttressed by the crafted pseudo labels for target samples. 
However, CDTrans has a distinct limitation: as the performance of cross-attention highly depends on the quality of pseudo labels, it becomes less effective when the domain gap becomes large. 
As shown in Fig.~\ref{fig:result_cover}, due to the significant gap between
the domain \textit{qdr} and the other domains, aligning distributions directly between them performs poorly. 

\begin{figure}
    \centering
    \includegraphics[width=0.8\linewidth]{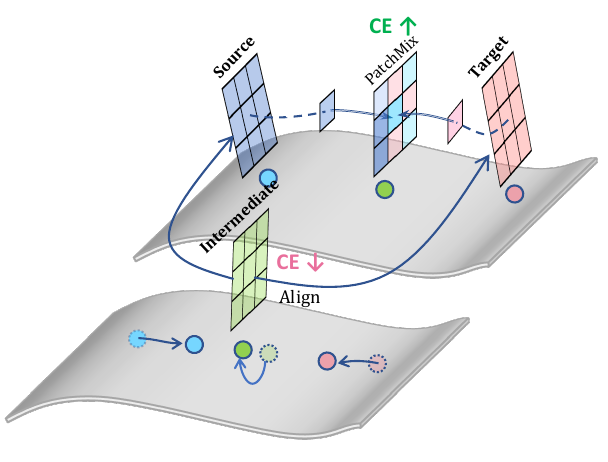}
    \captionsetup{font=small}
    \vspace{-8pt}
    \caption{PMTrans builds up the intermediate domain (\textcolor{green(pigment)}{green patches}) via a novel PatchMix module by learning to sample patches from the source (\textcolor{blue}{blue patches}) and target (\textcolor{pink}{pink patches}) domains. PatchMix tries to maximize the CE (\textcolor{green(pigment)}{$\uparrow$}) between the intermediate domain and source/target domain, while the feature extractor and classifier try to minimize it (\textcolor{red}{$\downarrow$}) for aligning domains.}
    \label{fig:cover_mix}
    \vspace{-15pt}
\end{figure}
In this paper, we probe a new problem for UDA: \textbf{\textit{how to smoothly bridge the source and target domains by constructing an intermediate domain with an effective ViT-based solution?}}
The intuition behind this is that, compared to direct aligning domains,  
alleviating the domain gap between the intermediate and source/target domain can facilitate domain alignment. 
Accordingly, we propose a novel and effective method, called \textbf{PMTrans} (PatchMix Transformer) to construct the intermediate representations. 
Overall, PMTrans interprets the process of domain alignment as a min-max cross entropy (CE) game with three players, \ie, the feature extractor, a classifier, and a \textbf{PatchMix} module, to find the Nash Equilibria. Importantly, the PatchMix module is proposed to effectively build up the intermediate domain,~\ie, probability distribution, by learning to sample patches from both domains with weights generated from a learnable Beta distribution based on the game-theoretical models~\cite{abs-2202-05352, bacsar1998dynamic, MazumdarRS20}, as shown in Fig.~\ref{fig:cover_mix}. That is, we aim to learn to mix patches from two domains to maximize the CE between the intermediate domain and source/target domain. Moreover, two semi-supervised mixup losses in the feature and label spaces are proposed to minimize the CE. Interestingly, \textbf{we conclude that the source and target domains are aligned if mixing the patch representations from two domains is equivalent to mixing the corresponding labels}. Therefore, the domain discrepancy can be measured based on the CE between the mixed patches and mixed labels. Eventually, the three players have no incentive to change their parameters to disturb CE, meaning the source and target domains are well aligned. Unlike existing mixup methods~\cite{ZhangCDL18, YunHCOYC19, UddinMSCB21}, our proposed PatchMix subtly learns to combine the element-wise global and local mixture by mixing patches from the source and target domains for ViT-based UDA. Moreover, we leverage the class activation mapping (CAM) from ViT to allocate the semantic information to re-weight the label of each patch, thus enabling us to obtain more domain-discriminative features. 

We conduct experiments on \textbf{four} benchmark datasets, including Office-31 \cite{SaenkoKFD10}, Office-Home \cite{VenkateswaraECP17}, VisDA-2017 \cite{abs-1710-06924}, and DomainNet \cite{PengBXHSW19}.
The results show that the performance of PMTrans significantly surpasses that of the ViT-based \cite{abs-2204-07683,abs-2109-06165,abs-2108-05988} and CNN-based SoTA methods~\cite{NaJCH21,LiXGLWL21,SharmaKC21} by \textbf{+3.6\%} on Office-Home, \textbf{+1.4\%} on Office-31, and \textbf{+7.2\%} on DomainNet (See Fig.~\ref{fig:result_cover}), respectively. 

Our main contributions are four-fold: (\textbf{I}) We propose a novel ViT-based UDA framework, PMTrans, to effectively bridge source and target domains by constructing the intermediate domain. (\textbf{II}) We propose PatchMix, a novel module to build up the intermediate domain via the game-theoretical models. (\textbf{III}) We propose two semi-supervised mixup losses in the feature and label spaces to reduce CE in the min-max CE game. (\textbf{IV}) Our PMTrans surpasses the prior methods by a large margin on three benchmark datasets.

\vspace{-8pt}
\section{Related Work}
\vspace{-3pt}
\noindent \textbf{Unsupervised Domain Adaptation.}
The prevailing UDA methods focus on domain alignment and learning discriminative domain-invariant features via metric learning, domain adversarial training, and optimal transport. Firstly, the metric learning-based methods aim to reduce the domain discrepancy by learning the domain-invariant feature representations using various metrics. For instance, some methods~\cite{LongC0J15, LongZ0J17, ZhuZW19, Kang0YH19} use the maximum mean discrepancy (MMD) loss to measure the divergence between different domains. In addition, the central moment discrepancy (CMD) loss~\cite{ZellingerGLNS17} and maximum density divergence (MDD) loss~\cite{LiCDZLS21} are also proposed to align the feature distributions. Secondly, the domain adversarial training methods learn the domain-invariant representations to encourage samples from different domains to be non-discriminative with respect to the domain labels via an adversarial loss~\cite{GaninUAGLLML16, XuZNLWTZ20, WuIE20}.
The third type of approach aims to minimize the cost transported from the source to the target distribution by finding an optimal coupling cost to mitigate the domain shift~\cite{CourtyFTR17, CourtyFHR17}. Unfortunately, these methods are not robust enough for the noisy pseudo target labels for accurate domain alignment.\textit{ Different from these mainstream UDA methods and ~\cite{AcunaLZF22}, we interpret the process of UDA as a min-max CE game and find the Nash Equilibria for domain alignment with an intermediate domain and a pure ViT-based solution.}

\noindent\textbf{Mixup.}
It is an effective data augmentation technique to prevent models from over-fitting by linearly interpolating two input data. Mixup types can be categorized into global mixup (e.g., Mixup \cite{ZhangCDL18} and Manifold-Mixup \cite{VermaLBNMLB19}) and local mixup (CutMix \cite{YunHCOYC19}, saliency-CutMix \cite{UddinMSCB21}, TransMix \cite{DBLP:journals/corr/abs-2111-09833}, and Tokenmix \cite{LiuLZ0022}). In CNN-based UDA tasks, several works \cite{XuZNLWTZ20, WuIE20, NaJCH21, Dai00T0D21} also use the mixup technique by linearly mixing the source and target domain data. \textit{In comparison, we unify the global and local mixup in our PMTrans framework by learning to form a mixed patch from the source/target patch as the input to ViT. We learn the hyperparameters of the mixup ratio for each patch, which is the \textbf{first} attempt to interpolate patches based on the distribution estimation. Accordingly, we propose PatchMix which effectively builds up the intermediate domain by sampling patches from both domains based on the game-theoretical models.}

\noindent\textbf{Transformer.} Vision Transformer(ViT) \cite{VaswaniSPUJGKP17} has recently been introduced to tackle the challenges in various vision tasks~\cite{CaronTMJMBJ21, LiuL00W0LG21}. Several works have leveraged ViT for the non-trivial UDA task. TVT~\cite{abs-2108-05988} proposes an adaptation module to capture domain data's transferable and discriminative features. SSRT \cite{abs-2204-07683} proposes a framework with a transformer backbone and a safe self-refinement strategy to handle the issue in case of a large domain gap. More recently, CDTrans \cite{abs-2109-06165} proposes a two-step framework that utilizes the cross-attention in ViT for direct feature alignment and pre-generated pseudo labels for the target samples. \textit{Differently, we probe to construct an intermediate domain to bridge the source and target domains for better domain alignment. Our PMTrans effectively interprets the process of domain alignment as
a min-max CE game, leading to a significant UDA performance enhancement}.

\begin{figure*}[t!]
    \centering
    
    \includegraphics[width=.85\textwidth]{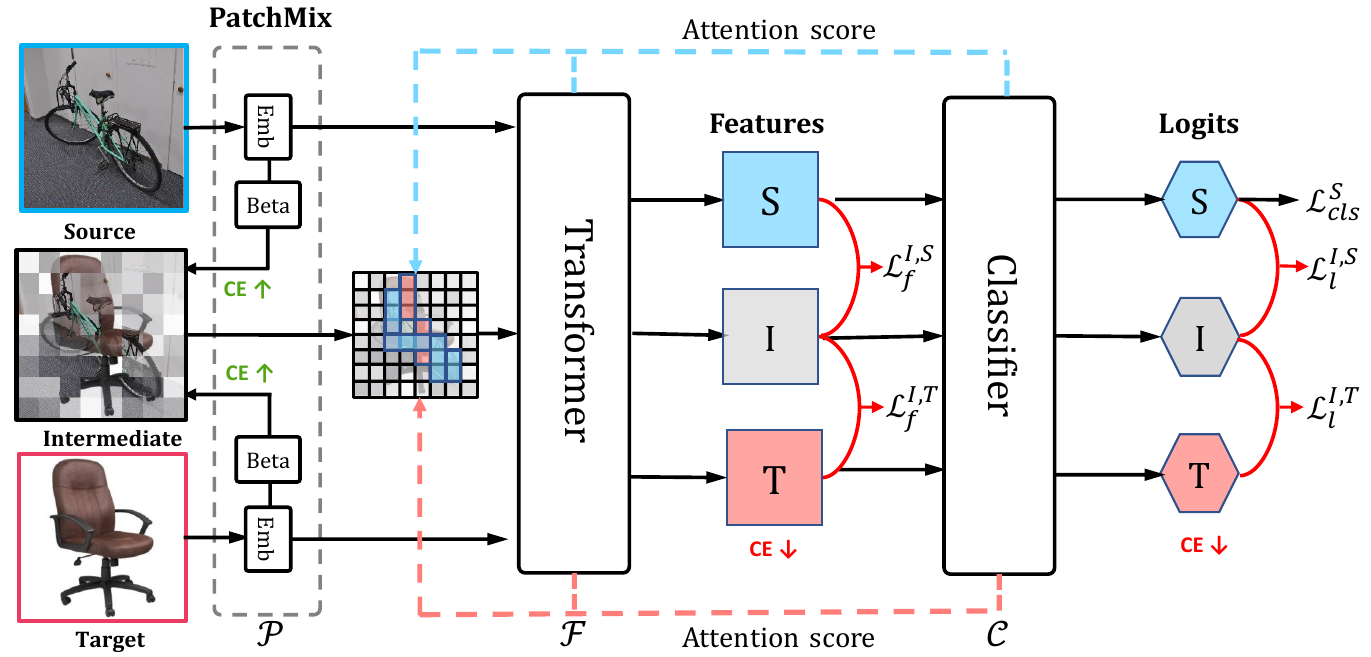}
    \captionsetup{font=small}
    \vspace{-10pt}
    \caption{Overview of the proposed PMTrans framework. It consists of three players:
    the PatchMix module empowered by a patch embedding (\textbf{Emb}) layer and a learnable Beta distribution (\textbf{Beta}), ViT encoder, and classifier. 
    }
    \label{fig:framework}
    \vspace{-15pt}
\end{figure*}

\vspace{-4pt}
\section{Methodology}\label{Methodology}
\vspace{-4pt}
In UDA, denote a labeled source set $\mathcal{D}_{s}=\left\{\left(\boldsymbol{x}^{s}_{i}, \boldsymbol{y}^{s}_{i}\right)\right\}_{i=1}^{n_{s}}$ with $i$-th sample $\boldsymbol{x}_{i}^{s} $ and its corresponding one-hot label $\boldsymbol{y}_{i}^{s}$ and an unlabeled target set $\mathcal{D}_{t}=$ $\left\{\boldsymbol{x}^{t}_{j}\right\}_{j=1}^{n_{t}}$ with $j$-th sample $\boldsymbol{x}_{j}^{t}$, $n_{s}$ and $n_{t}$ as the size of samples in the source and target domains, respectively. Note that the data in two domains are sampled from two different distributions, and we assume that the two domains share the same label space. Our goal is to address the significant domain divergence issue and smoothly transfer the knowledge from the source domain to the target domain. Firstly, we define and introduce PatchMix, and interpret the process of UDA as a min-max CE game. Secondly, we describe the proposed PMTrans which smoothly aligns the source and target domains by constructing an intermediate domain via a three-player game. 
\subsection{PMTrans: Theoretical Analysis}
\label{sub:PMTrans}
\subsubsection{PatchMix}

\begin{definition}
\label{Def: Mixup}
(PatchMix): Let $\mathcal{P}_{\lambda}$ be a linear interpolation operation on two pairs of randomly drawn samples $(\boldsymbol{x}^{s}, \boldsymbol{y}^{s})$ and $(\boldsymbol{x}^{t}, \boldsymbol{y}^{t})$. Then with $\lambda_k \sim \operatorname{Beta}(\beta, \gamma)$, it interpolates the $k$-th source patch $\boldsymbol{x}_k^{s}$ and target patch $\boldsymbol{x}_k^{t}$ to reconstruct a mixed representation with $n$ patches.
\begin{small}
\begin{equation}
\begin{aligned}
\label{eq:mixup_eq}
   &\boldsymbol{x}^{i}=\mathcal{P}_\lambda(\boldsymbol{x}^s, \boldsymbol{x}^t), ~\boldsymbol{x}^{i}_{k}=\lambda_{k} \odot \boldsymbol{x}_k^s + (1-\lambda_k)\odot\boldsymbol{x}_k^t,  \\ 
   &\boldsymbol{y}^{i}=\mathcal{P}_\lambda(\boldsymbol{y}^s, \boldsymbol{y}^t) = \frac{(\sum_{k=1}^{n}\lambda_{k})\boldsymbol{y}^s + (\sum_{k=1}^{n}(1-\lambda_{k}))\boldsymbol{y}^t}{n}. 
\end{aligned}
\end{equation}
\end{small}
\end{definition}


where $\boldsymbol{x}^{i}_{k}$ is the $k$-th patch of $\boldsymbol{x}^{i}$, and $\odot$ denotes multiplication. In Definition.~\ref{Def: Mixup}, each image $\boldsymbol{x}^{i}$ of the intermediate domain composes the sampled patches $\boldsymbol{x}_{k}$ from the source/target domain. Here, $\lambda_k \in [0,1]$ is the random mixing proportion that denotes the patch-level sampling weights. Furthermore, we calculate the image-level importance by aggregating patch weights $\sum_{k=1}^{n}(1-\lambda_{k})$, which is utilized to interpolate their labels. As a result, we mix both samples $(\boldsymbol{x}^s, \boldsymbol{y}^s)$ and $(\boldsymbol{x}^t, \boldsymbol{y}^t)$ to construct a new intermediate domain $\mathcal D_{i}=\left\{\left(\boldsymbol{x}^{i}_{l}, \boldsymbol{y}^{i}_{l}\right)\right\}_{l=1}^{n_{i}}$. 
To align the source and target domains, we need to evaluate the gap numerically. In detail, let $P_S$ and $P_T$ be the empirical distributions defined by $\mathcal{D}_{s}$ and $\mathcal{D}_{t}$, respectively. The domain divergence between source and target domains can be measured as 
{\setlength\abovedisplayskip{2pt}
\setlength\belowdisplayskip{2pt}
\begin{equation}
\label{eq:eq1}
\begin{aligned}
 D(P_S,P_T)
 =&\inf _{f \in \mathcal{F}, c \in \mathcal{C}, \mathcal{P}_{\lambda} \in \mathcal{P}} \underset{(\boldsymbol{x}^s, \boldsymbol{y}^s),\left(\boldsymbol{x}^{t}, \boldsymbol{y}^{t}\right)} {\mathbb{E}}\\ &\ell\left(c\left(\mathcal{P}_{\lambda}\left(f(\boldsymbol{x}^{s}), f\left(\boldsymbol{x}^{t}\right)\right)\right), \mathcal{P}_{\lambda}\left(\boldsymbol{y}^{s}, \boldsymbol{y}^{t}\right)\right),
\end{aligned}
\end{equation}}
where $\mathcal{F}$ denotes a set of encoding functions \ie, the feature extractor and $\mathcal{C}$ denotes a set of decoding functions \ie the classifier. Let $\mathcal{P}$ be the set of functions to generate the mixup ratio for building the intermediate domain. Then we can reformulate Eq. \ref{eq:eq1} as
{\setlength\abovedisplayskip{2pt}
\setlength\belowdisplayskip{2pt}
\begin{small}
\begin{equation}
\begin{aligned}
\label{eq:D}
  D\left(P_{S},P_{T}\right)
  =& \inf_{\boldsymbol{h}_{1}^{s}, \ldots, \boldsymbol{h}_{n_{s}}^{s} \in \mathcal{H}^{s},\boldsymbol{h}_{1}^{t}, \ldots, \boldsymbol{h}_{n_{t}}^{t} \in \mathcal{H}^{t}}\frac{1}{n_{s} \times n_{t}} \sum_{i}^{n_{s}}\sum_{j}^{n_{t}}\\
  &\left\{\inf _{c \in \mathcal{C}} \int_{0}^{1} \ell\left(f\left(\mathcal{P}_{\lambda}\left(\boldsymbol{h}_{i}^{s}, \boldsymbol{h}_{j}^{t}\right)\right), \mathcal{P}_{\lambda}\left(\boldsymbol{y}_{i}^{s}, \boldsymbol{y}_{j}^{t}\right)\right) p(\lambda) \mathrm{d} \lambda\right\},
\end{aligned}
\end{equation}
\end{small}}
where $\ell$ is CE loss, $\boldsymbol{h}_{i}^{s} = f(\boldsymbol{s}^{s}_{i})$ and $\boldsymbol{h}_{j}^{t} = f(\boldsymbol{x}^{t}_{j})$. Note $\mathcal{H}^{s}$ and $\mathcal{H}^{t}$ denote the representation spaces with dimensionality $\operatorname{dim}(\mathcal{H})$ for source and target domains, respectively. $\text {Let} f^{\star} \in \mathcal{F}, c^{\star} \in \mathcal{C}, \text{ and } {\mathcal{P}_{\lambda}}^{\star} \in \mathcal{P}$ be minimizers of Eq.~\ref{eq:eq1}.
\begin{theorem}
\label{theorem:Distribution estimation1}
(Domain Distribution Alignment with PatchMix): 
Let $d \in \mathbb{N}$ to represent the number of classes contained in three sets $\mathcal D_{s}$, $\mathcal D_{t}$, and $\mathcal D_{i}$. If $\operatorname{dim}(\mathcal{H}) \geq d-1$, ${\mathcal{P}_{\lambda}}’ \ell(c^{\star}(f^{\star}(\boldsymbol{x}_{i})),\boldsymbol{y}^{s})+ (1-{\mathcal{P}_{\lambda}}')\ell(c^{\star}(f^{\star}(\boldsymbol{x}_{i})),\boldsymbol{y}^{t})=0$, then $D\left(P_{S},P_{T}\right)=0$ and the corresponding minimizer $c^{\star}$ is a linear function from $\mathcal{H}$ to $\mathbb{R}^{d}$. Denote the scaled mixup ratio sampled from a learnable Beta distribution as $\mathcal{P}_{\lambda}’$.
\end{theorem}
Theorem.~\ref{theorem:Distribution estimation1} indicates that \textbf{the source and target domains are aligned if mixing the patches from two domains is equivalent to mixing the corresponding labels}. Therefore, minimizing the CE between the mixed patches and mixed labels can effectively facilitate domain alignment. \textit{For the proof of Theorem.~\ref{theorem:Distribution estimation1}, refer to the suppl. material.}

\vspace{-10pt}
\subsubsection{A Min-Max CE Game}
%
We interpret UDA as a min-max CE game among three players, namely the feature extractor ($\mathcal{F}$), classifier ($\mathcal{C}$), and PatchMix module ($\mathcal{P}$), as shown in Fig. \ref{fig:framework}. 
To specify each player's role, we define $\boldsymbol{\omega}_{\mathcal{F}} \in \Omega_{\mathcal{F}}$, $\boldsymbol{\omega}_{\mathcal{C}} \in \Omega_{\mathcal{C}}$, and $\boldsymbol{\omega}_{\mathcal{P}} \in \Omega_{\mathcal{P}}$ as the parameters of $\mathcal F$, $\mathcal C$, and $\mathcal P$, respectively. The joint domain is defined as $\Omega = \Omega_{\mathcal{F}} \times \Omega_{\mathcal{C}} \times \Omega_{\mathcal{P}}$ and their joint parameter set is defined as $\boldsymbol{\omega}=\{\boldsymbol{\omega}_{\mathcal{F}}, \boldsymbol{\omega}_{\mathcal{C}}, \boldsymbol{\omega}_{\mathcal{P}}\}$. 
%
Then we use the subscript $_{-m}$ to denote all other parameters/players except $m$, \eg, $\boldsymbol{\omega}_{-{\mathcal{C}}} = \{\boldsymbol{\omega}_{\mathcal{F}},\boldsymbol{\omega}_{\mathcal{P}}\}$. In our game, $m$-th player is endowed with a cost function $J_{m}$ and strives to reduce its cost, which contributes to the change of CE. Each player' cost function $J_{m}$ is represented as
{\setlength\abovedisplayskip{2pt}
\setlength\belowdisplayskip{2pt}
\begin{equation}
\small
\begin{split}\label{eq:objective1}
J_{{\mathcal{F}}}\left(\boldsymbol{\omega}_{\mathcal{F}}, \boldsymbol{\omega}_{-{\mathcal{F}}}\right) &:=\mathcal{L}_{cls}^{S}(\boldsymbol{\omega}_{\mathcal{F}}, \boldsymbol{\omega}_{\mathcal{C}})+\alpha {\text{CE}}_{s,i, t}(\boldsymbol{\omega}),\\ 
J_{{\mathcal{C}}}\left(\boldsymbol{\omega}_{\mathcal{C}}, \boldsymbol{\omega}_{-{\mathcal{C}}}\right) &:=\mathcal{L}_{cls}^{S}(\boldsymbol{\omega}_{\mathcal{F}}, \boldsymbol{\omega}_{\mathcal{C}})+\alpha {\text{CE}}_{s,i, t}(\boldsymbol{\omega}),\\ 
J_{{\mathcal{P}}}\left(\boldsymbol{\omega}_{{\mathcal{P}}}, \boldsymbol{\omega}_{-{\mathcal{P}}}\right) &:=-\alpha {\text{CE}}_{s,i, t}(\boldsymbol{\omega}),    
\end{split}
\end{equation}}
where $\alpha$ is the trade-off parameter, $\ell$ is the supervised classification loss for the source domain, and $\text{CE}_{s,i,t}(\boldsymbol{\omega})$ is the discrepancy between the intermediate domain and the source/target domain. The definitions of $\mathcal{L}_{cls}^{S}(\boldsymbol{\omega}_{\mathcal{F}}, \boldsymbol{\omega}_{\mathcal{C}})$ and $\text{CE}_{s,i,t}(\boldsymbol{\omega})$ are shown in Sec.~\ref{PMTransframework}. As illustrated in Eq.~\ref{eq:objective1}, the game is essentially a min-max process,~\ie, a competition for the player $\mathcal P$ against both players $\mathcal F$ and $\mathcal C$. Specifically, as depicted in Fig.~\ref{fig:framework}, $\mathcal P$ strives to diverge while $\mathcal{F}$ and $\mathcal C$ try to align domain distributions, which is a min-max process on CE. In this min-max CE game, each player behaves selfishly to reduce its cost function, and this competition will possibly end with a situation where no one has anything to gain by changing only one's strategy. This situation is called Nash Equilibrium (NE) in game theory. 


%
%
\begin{definition}
\label{Def:NE}
(Nash Equilibrium): The equilibrium states each player's strategy is the best response to other players. And a point ${\omega}^*\in \Omega$ is Nash Equilibrium if 
\begin{small}
\begin{equation}
    \begin{aligned}
    \forall \boldsymbol{\omega}_{m} \in {\Omega}_{i}, \forall m \in \{\mathcal{F}, \mathcal{C}, \mathcal{P}\}, s.t. J_m(\boldsymbol{\omega}_m^*, \boldsymbol{\omega}_{-m}^*) \leq J_m(\boldsymbol{\omega}_m, \boldsymbol{\omega}_{-m}^*). \nonumber
    \end{aligned}
\end{equation}
\end{small}
\end{definition}
Intuitively, in our case, NE means that no player has the incentive to change its own parameters, as there is no additional pay-off.

\subsection{The Proposed Framework}
\label{PMTransframework}
\noindent \textbf{Overview.}
Fig.~\ref{fig:framework} illustrates the framework of our proposed PMTrans, which consists of a ViT encoder, a classifier, and a PatchMix module. PatchMix module is utilized to maximize the CE between the intermediate domain and source/target domain, conversely, two semi-supervised mixup losses in the feature and label spaces are proposed to minimize CE. Finally, a three-player game containing feature extractor, classifier, and PatchMix module, minimizes and maximizes the CE for aligning distributions between the source and target domains.  

\noindent \textbf{PatchMix}.
As shown in Fig.~\ref{fig:framework}, the PatchMix module is proposed to construct the intermediate domain, buttressed by Definition.~\ref{Def: Mixup}. In detail, the patch embedding layer in PatchMix transforms input images from source/target domains into patches. And the ViT encoder aims to extract features from the patch sequences. The classifier maps the outputs of ViT encoder to make predictions, each of which is exploited to select the feature map to re-weight the patch sequences. The PatchMix with a learnable Beta distribution aims to maximize the CE between the intermediate and source/target domain, and is presented as follows.

When exploiting PatchMix to construct the intermediate domain, it is worth noting that not all patches have equal contributions for the label assignment. As Chen \etal\cite{DBLP:journals/corr/abs-2111-09833} observed, the mixed image has no valid objects due to the random process while there is still a response in the label space. 
To remedy this issue, we re-weight $\mathcal{P}_\lambda(\boldsymbol{y}^s, \boldsymbol{y}^t)$ in Definition. \ref{Def: Mixup} with the normalized attention score $a_{k}$. For the implementation details of attention scores, refer to \textit{the suppl. material}. The re-scaled $\mathcal{P}_\lambda(\boldsymbol{y}^s, \boldsymbol{y}^t)$ is defined as 
{\setlength\abovedisplayskip{3pt}
\setlength\belowdisplayskip{3pt}
\begin{equation}
    \begin{aligned}
    \mathcal{P}_\lambda(\boldsymbol{y}^s, \boldsymbol{y}^t) = \lambda^s\boldsymbol{y}^s + \lambda^t\boldsymbol{y}^t, \nonumber
    \end{aligned}
\end{equation} }
where
{\setlength\abovedisplayskip{5pt}
\setlength\belowdisplayskip{3pt}
\begin{equation}
\small
    \begin{aligned}
\lambda^s &= \frac{\sum_{k=1}^{n} \lambda_{k} {a}_{k}^{s}}{\sum_{k=1}^{n} \lambda_{k} {a}_{k}^{s}+\sum_{k=1}^{n}(1-\lambda_{k}){a}_{k}^{t}}
, \\
\lambda^t &= \frac{\sum_{k=1}^{n}(1-\lambda_{k}) {a}_{k}^{t}}{\sum_{k=1}^{n} \lambda_{k} {a}_{k}^{s}+\sum_{k=1}^{n}(1-\lambda_{k}){a}_{k}^{t}}.\nonumber
    \end{aligned}
\end{equation}}
\begin{figure}
    \centering
    \includegraphics[width=\linewidth]{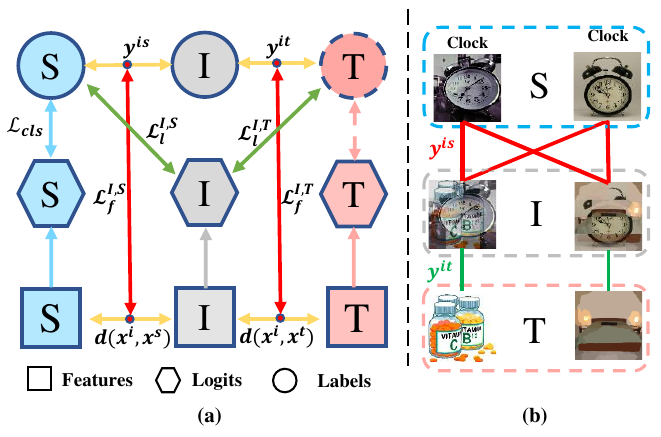}
    \captionsetup{font=small}
    \vspace{-15pt}
    \caption{(a) The illustration of two proposed semi-supervised losses. (b) Label similarity $\boldsymbol{y}^{is}$ and $\boldsymbol{y}^{it}$. Better viewed in color.}
    \label{fig:align-feature}
    \vspace{-15pt}
\end{figure}
\noindent\textbf{Semi-supervised mixup loss}. 
As PatchMix tries to maximize the CE between the intermediate domain and source/target domain, we now need to find a way to minimize the CE in the game. Intuitively, we propose two semi-supervised mixup losses in the feature and label spaces to minimize the discrepancy between features of mixing patches and corresponding mixing labels based on Theorem.~\ref{theorem:Distribution estimation1}. The objective of the proposed PMTrans is show in Fig. \ref{fig:align-feature} \textcolor{red}{(a)} and consists of a classification loss of source data and two semi-supervised mixup losses.

\textbf{ 1) Label space:} As introduced in Theorem.~\ref{theorem:Distribution estimation1}, we apply a supervised mixup loss in the label space to measure the domain divergence based on the CE loss between the mixing logits and corresponding mixing labels (See \textcolor{green(pigment)}{green arrow} in Fig. \ref{fig:align-feature} \textcolor{red}{(a)}). 
{\setlength\abovedisplayskip{3pt}
\setlength\belowdisplayskip{3pt}
\begin{equation}
\small
    \begin{aligned}
 \mathcal{L}_{l}^{I,S}(\boldsymbol{\omega}) &= \mathbb{E}_{\left(\boldsymbol{x}^{i}, \boldsymbol{y}^{i}\right) \sim D^{i}} \lambda^{s} \ell\left(\mathcal{C}\left(\mathcal {F}\left(\boldsymbol{x}^{i}\right)\right), \boldsymbol{y}^{s}\right),\\
\mathcal{L}_{l}^{I,T}(\boldsymbol{\omega}) &= \mathbb{E}_{\left(\boldsymbol{x}^{i}, \boldsymbol{y}^{i}\right) \sim D^{i}} \lambda^{t} \ell\left(\mathcal{C}\left(\mathcal {F}\left(\boldsymbol{x}^{i}\right)\right), \hat{\boldsymbol{y}^{t}}\right), \nonumber
    \end{aligned}
\end{equation}}

where $\hat{\boldsymbol{y}^{t}}$ is the pseudo label for target data. For convenience, we utilize the method, commonly used in~\cite{LiangHF20, LiangHF21}, to generate pseudo labels $ \hat{\boldsymbol{y}^{t}}$ for samples via $k$-means cluster. 

\textbf{2) Feature space:} Nonetheless, the supervised loss alone in the label space is
not sufficient to diminish the domain divergence due to the less reliable pseudo labels of the target data. 
Therefore, we further  \textbf{propose to minimize the discrepancy between the similarity of the features and the similarity of labels in the feature space} for aligning the intermediate and source/target domain without the supervised information of the target domain. The experimental results in Tab.~\ref{tab:semi1} validate its effectiveness.

Specifically, we first compute the cosine similarity between the intermediate domain and source/target domain in the feature space. The feature similarity is defined as
{\setlength\abovedisplayskip{3pt}
\setlength\belowdisplayskip{3pt}
\begin{equation}
    d(\boldsymbol{x}^{i}, \boldsymbol{x}^{s}) = cos(\mathcal {F}\left(\boldsymbol{x}^{i}\right), \mathcal {F}\left(\boldsymbol{x}^{s}\right)), \nonumber
\end{equation}}
where $cos$ denotes the cosine similarity. 
As shown in Fig.~\ref{fig:align-feature}\textcolor{red}{(b)}, for the source domain, we exploit the ground-truth to calculate the label similarity, $y^{is} =y^s({y^s})^{\intercal}$, as a binary matrix to represent whether samples share the same labels. For example, the intermediate image is constructed by sampling patches from the source image,~\eg, \textit{clock}; therefore, the label similarity is set as 1 if it is calculated between the intermediate image and the source class \textit{`clock'}, otherwise, it is set as 0.
%
Then, we utilize the CE to measure the domain discrepancy based on the difference between the feature similarity and label similarity. The supervised mixup loss in the feature space (See \textcolor{red}{red arrow} in Fig. \ref{fig:align-feature} \textcolor{red}{(a)}) is formulated as
{\setlength\abovedisplayskip{3pt}
\setlength\belowdisplayskip{3pt}
\begin{small}
\begin{equation}
    \begin{aligned}
    \mathcal{L}_{f}^{I,S}(\boldsymbol{\omega}_{\mathcal{F}}, \boldsymbol{\omega}_{\mathcal{P}}) = \mathbb{E}_{\left(\boldsymbol{x}^{i}, \boldsymbol{y}^{i}\right) \sim D^{i}} \lambda^{s}\ell\left(d(\boldsymbol{x}^{i},\boldsymbol{x}^{s}), \boldsymbol{y}^{is}\right). \nonumber
    \end{aligned}
\end{equation}
\end{small}
}
Moreover, for the intermediate and target domains, due to lack of supervision, we utilize identity matrix $y^{it}$ as the label similarity. For example, in Fig.~\ref{fig:align-feature} \textcolor{red}{(b)}, as the intermediate image is built by sampling patches from the target image, \eg, bottle; therefore, the label similarity between the intermediate image and the corresponding target image is set as 1 and vice versa. 
To measure the divergence between the intermediate and target domains in the feature space, we propose an unsupervised mixup loss as
{\setlength\abovedisplayskip{3pt}
\setlength\belowdisplayskip{3pt}
\begin{small}
\begin{equation}
\small
    \begin{aligned}
   \mathcal{L}_{f}^{I,T}(\boldsymbol{\omega}_{\mathcal{F}}, \boldsymbol{\omega}_{\mathcal{P}}) = \mathbb{E}_{\left(\boldsymbol{x}^{i}, \boldsymbol{y}^{i}\right) \sim D^{i}} \lambda^{t}\ell\left(d(\boldsymbol{x}^{i}, \boldsymbol{x}^{t}), \boldsymbol{y}^{it}\right), \nonumber
    \end{aligned}
\end{equation}
\end{small}}

Finally, the two semi-supervised mixup losses in the feature and label spaces are formulated as
{\setlength\abovedisplayskip{3pt}
\setlength\belowdisplayskip{3pt}
\begin{equation}
\small
    \begin{aligned}
 \mathcal{L}_{f}(\boldsymbol{\omega}_{\mathcal{F}}, \boldsymbol{\omega}_{\mathcal{P}}) &= \mathcal{L}_{f}^{I,S}(\boldsymbol{\omega}_{\mathcal{F}}, \boldsymbol{\omega}_{{\mathcal{P}}}) + \mathcal{L}_{f}^{I,T}(\boldsymbol{\omega}_{\mathcal{F}}, \boldsymbol{\omega}_{\mathcal{P}}),\\
\mathcal{L}_{l}(\boldsymbol{\omega}) &= \mathcal{L}_{l}^{I,S}(\boldsymbol{\omega}) + \mathcal{L}_{l}^{I,T}(\boldsymbol{\omega}). \nonumber
    \end{aligned}
\end{equation}}
Moreover, the classification loss is applied to the labeled source domain data (See \textcolor{blue_light}{blue arrow} in Fig. \ref{fig:align-feature} \textcolor{red}{(a)}) and is formulated as 
{\setlength\abovedisplayskip{3pt}
\setlength\belowdisplayskip{3pt}
\begin{equation}
\small
    \begin{aligned}
\mathcal{L}_{cls}^{S}(\boldsymbol{\omega}_{\mathcal{F}}, \boldsymbol{\omega}_{\mathcal{C}})=\mathbb{E}_{\left(\boldsymbol{x}^{s}, \boldsymbol{y}^{s}\right) \sim D^{s}} \ell\left(\mathcal{C}\left(\mathcal{F}\left(\boldsymbol{x}^{s}\right)\right), \boldsymbol{y}^{s}\right).\nonumber
    \end{aligned}
\end{equation}}
\begin{table*}[t]
\centering
\resizebox{0.9\linewidth}{!}{ 
\begin{tabular}{c|l|llllllllllllll}
\toprule
Method &&A$\to$ C& A$\to$ P & A $\to$ R & C $\to$ A &C $\to$ P &C $\to$ R &P$\to$ A& P$\to$ C & P$ \to$ R & R$ \to$ A &R$ \to$ C &R$ \to$ P& Avg     \\
\hline
\hline
ResNet-50 &\multirow{5}{*}{\rotatebox{90}{ResNet}}& 44.9& 66.3 &74.3 &51.8& 61.9 &63.6& 52.4 &39.1& 71.2& 63.8& 45.9& 77.2 &\colorbox{lightgray}{59.4}   \\
MCD &&48.9& 68.3& 74.6& 61.3& 67.6& 68.8& 57.0& 47.1 &75.1& 69.1& 52.2& 79.6& \colorbox{lightgray}{64.1}\\
MDD &&54.9& 73.7& 77.8& 60.0& 71.4& 71.8& 61.2& 53.6& 78.1& 72.5& 60.2& 82.3& \colorbox{lightgray}{68.1}\\
BNM&&56.7 &77.5& 81.0& 67.3& 76.3& 77.1& 65.3& 55.1& 82.0 &73.6 &57.0& 84.3& \colorbox{lightgray}{71.1}\\
FixBi && 58.1& 77.3& 80.4 &67.7& 79.5& 78.1& 65.8& 57.9& 81.7& 76.4& 62.9& 86.7& \colorbox{lightgray}{72.7}  \\
\midrule
TVT & \multirow{7}{*}{\rotatebox{90}{ViT}}&   74.9&86.8&89.5&82.8&88.0&88.3&79.8&71.9&90.1&85.5&74.6&90.6&\colorbox{lightgray}{83.6}\\
Deit-based&&61.8 &79.5& 84.3& 75.4 &78.8& 81.2& 72.8 &55.7& 84.4& 78.3& 59.3& 86.0 &\colorbox{lightgray}{74.8} \\
CDTrans-Deit&  & 68.8&85.0&86.9&81.5&87.1&87.3&79.6&63.3&88.2&82.0&66.0&90.6&\colorbox{lightgray}{80.5}   \\
\textbf{PMTrans-Deit} &&71.8& 87.3 &88.3 &83.0 &87.7 & 87.8 &78.5 &67.4 &89.3 &81.7 &70.7 & 92.0 &\colorbox{lightgray}{82.1}\\
ViT-based &&67.0 &85.7 &88.1 &80.1 &84.1 &86.7 &79.5 &67.0 &89.4 &83.6 &70.2 &91.2 & \colorbox{lightgray}{81.1}\\
SSRT-ViT&&75.2& 89.0& 91.1& 85.1 &88.3& 89.9& 85.0& 74.2 &91.2& 85.7& 78.6 &91.8& \colorbox{lightgray}{85.4}\\
\textbf{PMTrans-ViT} &&81.2&91.6&92.4&\textbf{88.9}&91.6&93.0&\textbf{88.5}&80.0&\textbf{93.4}&\textbf{89.5}&\textbf{82.4}&94.5&\colorbox{lightgray}{88.9}\\
\midrule
Swin-based&\multirow{2}{*}{\rotatebox{90}{Swin}}&72.7   &87.1   &90.6   &84.3   &87.3   &89.3   &80.6 &68.6 &90.3&84.8&69.4&91.3& \colorbox{lightgray}{83.6}  \\
\textbf{PMTrans-Swin} && \textbf{81.3}&\textbf{92.9}&\textbf{92.8}&88.4&\textbf{93.4}&\textbf{93.2}&87.9&\textbf{80.4}&93.0&89.0&80.9&\textbf{94.8}&\colorbox{lightgray}{\textbf{89.0}}\\
\bottomrule
\end{tabular}}
\vspace{-8pt}
\caption{Comparison with SoTA methods on Office-Home. The best performance is marked as \textbf{bold}.}
\captionsetup{font=small}
\label{tab:officeHome_Res}
\vspace{-10pt}
\end{table*}
\begin{table}[t]
\centering
\setlength{\tabcolsep}{1.5mm}
\captionsetup{font=small}
\resizebox{1\linewidth}{!}{
\begin{tabular}{c|l|lllllll}
\toprule
Method  &&A $\to$ W & D$\to$ W & W$\to$ D & A$\to$ D &D$\to$ A &W$\to$ A & \colorbox{lightgray}{Avg}  \\
\hline
\hline
ResNet-50 &\multirow{5}{*}{\rotatebox{90}{ResNet}} &  68.9& 68.4 &62.5& 96.7& 60.7& 99.3& \colorbox{lightgray}{76.1}   \\
BNM &&91.5&98.5&\textbf{100.0}&90.3&70.9&71.6&\colorbox{lightgray}{87.1}\\
MDD&&94.5& 98.4& \textbf{100.0}& 93.5& 74.6 &72.2& \colorbox{lightgray}{88.9}\\
SCDA&&94.2 &98.7& 99.8& 95.2& 75.7& 76.2& \colorbox{lightgray}{90.0}\\
FixBi &  &  96.1& 99.3& \textbf{100.0}& 95.0&78.7& 79.4& \colorbox{lightgray}{91.4}  \\ 
\midrule
TVT &\multirow{7}{*}{\rotatebox{90}{ViT}}&96.4& 99.4& \textbf{100.0}& 96.4& 84.9& 86.0&\colorbox{lightgray}{93.9}   \\
Deit-based &     &89.2& 98.9& \textbf{100.0}& 88.7& 80.1& 79.8& \colorbox{lightgray}{89.5}\\
CDTrans-Deit& & 96.7 &99.0& \textbf{100.0}& 97.0& 81.1& 81.9 &\colorbox{lightgray}{92.6}  \\
\textbf{PMTrans-Deit}&&99.0&99.4&\textbf{100.0}&96.5&81.4&82.1&\colorbox{lightgray}{93.1}\\
ViT-based       &&91.2& 99.2& \textbf{100.0}& 90.4& 81.1& 80.6& \colorbox{lightgray}{91.1}  \\
SSRT-ViT&&97.7& 99.2 &\textbf{100.0} &98.6& 83.5& 82.2& \colorbox{lightgray}{93.5}\\
\textbf{PMTrans-ViT}&& 99.1&\textbf{99.6}&\textbf{100.0}&99.4&85.7&86.3&\colorbox{lightgray}{95.0}\\
\midrule
Swin-based &\multirow{2}{*}{\rotatebox{90}{Swin}}&97.0   &99.2   &\textbf{100.0}   &95.8   &82.4   &81.8   &\colorbox{lightgray}{92.7}   \\
\textbf{PMTrans-Swin} & &\textbf{99.5}   &99.4  &\textbf{100.0}   &\textbf{99.8}   &\textbf{86.7}   &\textbf{86.5}   &\colorbox{lightgray}{\textbf{95.3}}  \\ 
\bottomrule
\end{tabular}}
\vspace{-8pt}
\caption{Comparison with SoTA methods on Office-31. The best performance is marked as \textbf{bold}.}
\label{tab:office31_res}
\vspace{-10pt}
\end{table}
\noindent\textbf{A Three-Player Game}.
%
Finally, the min-max CE game aims to align distributions in the feature and label spaces. The total CE between the intermediate domain and source/target domain is 
{\setlength\abovedisplayskip{3pt}
\setlength\belowdisplayskip{3pt}
\begin{equation}
{\text{CE}}_{s,i,t}(\boldsymbol{\omega}) =  \mathcal{L}_{f}(\boldsymbol{\omega}_{\mathcal{F}}, \boldsymbol{\omega}_{\mathcal{P}})+\mathcal{L}_{l}(\boldsymbol{\omega}).  \nonumber  
\end{equation}}

We adopt the \textit{random} mixup-ratio from a learnable Beta distribution in our PatchMix module to maximize the CE between the intermediate domain and source/target domain. Moreover, the feature extractor and classifier have the same objective to minimize the CE between the intermediate domain and source/target domain.
Therefore, the total objective of PMTrans is achieved by reformulating Eq.~\ref{eq:objective1} as
\[
J\left(\boldsymbol{\omega}\right) :=\mathcal{L}_{cls}^{S}(\boldsymbol{\omega}_{\mathcal{F}}, \boldsymbol{\omega}_{\mathcal{C}})+\alpha {\text{CE}}_{s,i,t}(\boldsymbol{\omega}),
\]
where $\alpha$ is trade-off parameter. 
After optimizing the objective, the PatchMix module with the ideal Beta distribution
will not maximize the CE anymore. Meanwhile, the feature extractor and classifier have no incentive to change their parameters to minimize the CE. Finally, the discrepancy between the intermediate domain and  source/target domain is nearly zero, further indicating that the source and target domains are well aligned.
\section{Experiments}
\subsection{Datasets and implementation}
\noindent\textbf{Datasets.} To evaluate the proposed method, we conduct experiments on four popular UDA benchmarks, including Office-Home \cite{VenkateswaraECP17}, Office-31 \cite{SaenkoKFD10}, VisDA-2017 \cite{abs-1710-06924}, and DomainNet \cite{PengBXHSW19}.
\textit{The details of the datasets and transfer tasks on these datasets can be found in the suppl. material}.

\noindent\textbf{Implementation.} In all experiments, we use the Swin-based transformer~\cite{Liu2021SwinTH}  pre-trained on ImageNet \cite{DengDSLL009} as the backbone for our PMTrans. The base learning rate is $5e^{-6}$ with a batch size of 32, and we train models by 50 epochs. For VisDA-2017, we use a lower learning rate $1e^{-6}$. We adopt AdamW \cite{LoshchilovH19} with a momentum of $0.9$, and a weight decay of $0.05$ as the optimizer. Furthermore, for fine-tuning purposes, we set the classifier (MLP) with a higher learning rate $1e^{-5}$ for our main tasks and learn the trade-off parameter adaptively. For a fair comparison with prior works, we also conduct experiments with the same backbone \textbf{Deit-based} \cite{DBLP:journals/corr/abs-2012-12877} as CDTrans \cite{abs-2109-06165}, and \textbf{ViT-based} \cite{Dosovitskiy2021AnII} as SSRT \cite{abs-2204-07683} on Office-31, Office-Home, and VisDA-2017. These two studies are trained for 60 and 100 epochs separately. 


\subsection{Results}

We compare PMTrans with the SoTA methods, including ResNet-based and ViT-based methods. The ResNet-based methods are FixBi \cite{NaJCH21}, MCD \cite{SaitoWUH18}, SWD \cite{LeeBBU19}, SCDA \cite{0008XLLLQL21}, BNM \cite{CuiWZLH020}, and MDD \cite{0002LLJ19}. The ViT-based methods are SSRT \cite{abs-2204-07683}, CDTrans \cite{abs-2109-06165}, and TVT \cite{abs-2108-05988}. 

 For the ResNet-based methods, we utilize ResNet-50 as the backbone for the Office-Home, Office-31, and DomainNet datasets, and we adopt ResNet-101 for VisDA-2017 dataset. Note that each backbone is trained with the source data only and then tested with the target data. 

\begin{table*}[t]
\centering
\captionsetup{font=small}
\resizebox{0.85\linewidth}{!}{ 
\begin{tabular}{c|l|lllllllllllll}
\toprule
Method &&plane& bcycl& bus &car& horse& knife& mcycl& person& plant &sktbrd &train& truck& Avg     \\
\hline
\hline
ResNet-50&\multirow{7}{*}{\rotatebox{90}{ResNet}}  & 55.1 &53.3& 61.9& 59.1& 80.6 &17.9 &79.7& 31.2& 81.0 &26.5 &73.5& 8.5& \colorbox{lightgray}{52.4} \\
BNM &&89.6 &61.5 &76.9 &55.0& 89.3& 69.1 &81.3 &65.5& 90.0 &47.3& 89.1 &30.1& \colorbox{lightgray}{70.4}\\
MCD& &  87.0 &60.9 &83.7& 64.0& 88.9& 79.6& 84.7& 76.9& 88.6& 40.3& 83.0& 25.8 &\colorbox{lightgray}{71.9}\\
SWD&&90.8 &82.5 &81.7 &70.5& 91.7 &69.5& 86.3& 77.5 &87.4& 63.6& 85.6 &29.2 &\colorbox{lightgray}{76.4}\\
FixBi && 96.1 &87.8& 90.5& 90.3 &96.8& 95.3 &92.8& 88.7& 97.2 &94.2& 90.9 &25.7& \colorbox{lightgray}{87.2}  \\ 
\midrule
TVT&\multirow{7}{*}{\rotatebox{90}{ViT}}& 82.9 &85.6 &77.5 &60.5& 93.6& 98.2& 89.4& 76.4& 93.6& 92.0& 91.7& 55.7&\colorbox{lightgray}{83.1}\\
Deit-based & &  98.2 &73.0 &82.5& 62.0 &97.3& 63.5& 96.5& 29.8&68.7 &86.7 &96.7&23.6&\colorbox{lightgray}{73.2} \\
CDTrans-Deit&&97.1& 90.5& 82.4& 77.5& 96.6&96.1 &93.6 &\textbf{88.6}&\textbf{97.9}&86.9 &90.3&\textbf{62.8}&\colorbox{lightgray}{88.4}\\
\textbf{PMTrans-Deit}&&98.2& 92.2 &88.1 &77.0 &97.4&95.8 &94.0&72.1 &97.1 &95.2 &94.6&51.0 &\colorbox{lightgray}{87.7}\\
ViT-based & &99.1& 60.7& 70.1& 82.7& 96.5 &73.1& 97.1 &19.7 &64.5& 94.7 &97.2& 15.4&\colorbox{lightgray}{72.6}\\
SSRT-ViT&&98.9& 87.6 &\textbf{89.1}&\textbf{84.8}& 98.3& \textbf{98.7}& 96.3&81.1& 94.8 &97.9& 94.5 &43.1& \colorbox{lightgray}{\textbf{88.8}}\\
\textbf{PMTrans-ViT}&& 98.9 & \textbf{93.7} &84.5 &73.3 &\textbf{99.0} &98.0 &96.2 &67.8 &94.2 &\textbf{98.4} &96.6 &49.0 & \colorbox{lightgray}{87.5}\\
\midrule
Swin-based &\multirow{2}{*}{\rotatebox{90}{Swin}}        &99.3   &  63.4 &  85.9 & 68.9  &  95.1 & 79.6  &\textbf{ 97.1}&29.0&81.4&94.2&\textbf{97.7}&29.6&\colorbox{lightgray}{76.8}  \\
\textbf{PMTrans-Swin}&& \textbf{99.4}                        & 88.3  & 88.1  &78.9   & 98.8  &98.3 &95.8&70.3   &94.6&98.3&96.3&48.5& \colorbox{lightgray}{88.0}  \\ 
\bottomrule
\end{tabular}}
\vspace{-8pt}
\caption{Comparison with SoTA methods on VisDA-2017. The best performance is marked as \textbf{bold}.}
\label{tab:VisDA1}
\end{table*}
\begin{table*}
\centering
\vspace{-5pt}
\captionsetup{font=small}
\resizebox{\linewidth}{!}{ 
\begin{tabular}{c|lllllll||c|lllllll||c|llllllll}
\toprule
MCD                 & clp& inf& pnt& qdr& rel& skt& Avg &SWD & clp& inf& pnt& qdr& rel& skt& Avg&BNM & clp& inf& pnt& qdr& rel& skt& Avg  \\
\hline
\hline
clp                      &  - &15.4 &25.5& 3.3& 44.6& 31.2 &24.0  &clp&  -& 14.7& 31.9& 10.1& 45.3& 36.5& 27.7&clp                      &  -& 12.1 &33.1& 6.2& 50.8 &40.2& 28.5  \\
inf                     &  24.1& -& 24.0 &1.6 &35.2& 19.7& 20.9 &inf &   22.9& -& 24.2& 2.5& 33.2& 21.3& 20.0&inf                     &   26.6 &- &28.5& 2.4 &38.5& 18.1& 22.8\\
pnt                      & 31.1& 14.8& -& 1.7& 48.1& 22.8& 23.7& pnt &  33.6 &15.3 &- &4.4 &46.1& 30.7 &26.0&pnt                      &  39.9& 12.2& - &3.4& 54.5& 36.2 &29.2\\
qdr                      &  8.5& 2.1 &4.6& -& 7.9& 7.1& 6.0& qdr &  15.5 &2.2& 6.4& -& 11.1& 10.2 &9.1&qdr                      &  17.8 &1.0 &3.6& - &9.2& 8.3& 8.0\\
rel                      &  39.4& 17.8& 41.2 &1.5& -& 25.2& 25.0&real& 41.2& 18.1& 44.2& 4.6& -& 31.6& 27.9 &rel                      & 48.6 &13.2& 49.7& 3.6& - &33.9& 29.8 \\
skt                     &   37.3& 12.6& 27.2 &4.1 &34.5& - &23.1& skt & 44.2& 15.2&37.3 &10.3& 44.7& -& 30.3 &skt                     &   54.9 &12.8 &42.3 &5.4 &51.3& - &33.3  \\ 
Avg                      &  28.1 &12.5& 24.5& 2.4& 34.1& 21.2 &\colorbox{lightgray}{20.5}& Avg & 31.5& 13.1& 28.8 &6.4& 36.1& 26.1& \colorbox{lightgray}{23.6}&Avg                      & 37.6& 10.3& 31.4& 4.2& 40.9& 27.3 & \colorbox{lightgray}{25.3} \\
\hline
\hline
CGDM & clp& inf& pnt& qdr& rel& skt& Avg&MDD& clp& inf& pnt& qdr& rel& skt& Avg&SCDA & clp& inf& pnt& qdr& rel& skt& Avg  \\
\hline
\hline
clp& -& 16.9& 35.3& 10.8& 53.5 &36.9& 30.7 & clp& -& 20.5& 40.7& 6.2& 52.5& 42.1& 32.4&clp& -& 18.6& 39.3& 5.1& 55.0& 44.1& 32.4\\
inf &27.8 &- &28.2& 4.4 &48.2& 22.5& 26.2& inf& 33.0& - &33.8& 2.6 &46.2 &24.5 &28.0&inf &29.6& - &34.0 &1.4 &46.3& 25.4& 27.3 \\ 
pnt& 37.7& 14.5& - &4.6& 59.4& 33.5& 30.0  &pnt &43.7& 20.4& - &2.8& 51.2 &41.7& 32.0&pnt& 44.1& 19.0& -& 2.6& 56.2& 42.0& 32.8 \\
qdr& 14.9 &1.5& 6.2& - &10.9& 10.2& 8.7  &qdr& 18.4& 3.0& 8.1 &- &12.9 &11.8& 10.8&qdr& 30.0 &4.9 &15.0 &- &25.4 &19.8 &19.0\\
rel &49.4& 20.8& 47.2& 4.8& - &38.2& 32.0 & rel&52.8& 21.6& 47.8& 4.2 &-& 41.2& 33.5&rel &54.0 &22.5& 51.9 &2.3& - &42.5& 34.6\\
skt& 50.1&16.5& 43.7& 11.1& 55.6& - &35.4 & skt& 54.3 &17.5 &43.1 &5.7 &54.2& - &35.0&skt& 55.6 &18.5 &44.7 &6.4 &53.2 &- &35.7\\ 
Avg& 36.0& 14.0& 32.1& 7.1 &45.5& 28.3& \colorbox{lightgray}{27.2} &Avg& 40.4 &16.6& 34.7& 4.3& 43.4 &32.3& \colorbox{lightgray}{28.6}&Avg& 42.6& 16.7& 37.0& 3.6 &47.2& 34.8& \colorbox{lightgray}{30.3}\\
\hline
\hline
CDTrans& clp& inf& pnt& qdr& rel& skt& Avg&SSRT& clp& inf& pnt& qdr& rel& skt& Avg&\textbf{PMTrans}& clp& inf& pnt& qdr& rel& skt& Avg\\
\hline
\hline
clp  &  - &29.4& 57.2& 26.0& 72.6& 58.1 &48.7&clp&-& 33.8 &60.2&19.4& 75.8 &59.8 &49.8&    clp & - & 34.2 &62.7 &32.5 &79.3 &63.7 & 54.5  \\
inf  &   57.0 &- &54.4 &12.8& 69.5& 48.4 &48.4&inf&55.5 &- &54.0 &9.0 &68.2 &44.7& 46.3    &inf& 67.4&- &61.1 & 22.2 & 78.0 & 57.6&57.3 \\ 
pnt  &  62.9& 27.4& -& 15.8& 72.1& 53.9& 46.4&pnt&61.7& 28.5& -& 8.4& 71.4& 55.2& 45.0&     pnt&69.7&33.5&-&23.9&79.8&61.2&53.6 \\
qdr &  44.6& 8.9 &29.0 &- &42.6 &28.5 &30.7&qdr&42.5 &8.8 &24.2& - &37.6 &33.6 &29.3 &      qdr&54.6&17.4&38.9&-&49.5  &41.0&40.3\\
rel &  66.2& 31.0& 61.5& 16.2& -& 52.9& 45.6&rel&69.9 &37.1& 66.0& 10.1& - &58.9& 48.4&       rel&74.1&35.3&70.0&25.4&-&61.1&53.2 \\
skt  &  69.0& 29.6 &59.0& 27.2 &72.5& -& 51.5&skt&70.6 &32.8 &62.2 &21.7 &73.2 &- &52.1&       skt&73.8&33.0&62.6&30.9&77.5&-& 55.6\\ 
Avg  & 59.9 &25.3 &52.2& 19.6& 65.9 &48.4& \colorbox{lightgray}{45.2}&Avg&60.0& 28.2& 53.3& 13.7& 65.3 &50.4& \colorbox{lightgray}{45.2}&     Avg&67.9&30.7&59.1&27.0&72.8&56.9&\colorbox{lightgray}{\textbf{52.4}}  \\
\bottomrule
\end{tabular}}
\vspace{-8pt}
\caption{Comparison with SoTA methods on DomainNet. The best performance is marked as \textbf{bold}.}
\vspace{-10pt}
\label{tab:domainnet}
\end{table*}


\noindent\textbf{Results on Office-Home.} Tab.~\ref{tab:officeHome_Res} shows the quantitative results of methods using different backbones. As expected, our PMTrans framework achieves noticeable performance gains and surpasses TVT, SSRT, and CDTrans by a large margin. Importantly, our PMTrans achieves an improvement more than \textbf{5.4\%} accuracy over the Swin backbone and yields \textbf{89.0\%} accuracy. Interestingly, our proposed PMTrans can decrease the domain divergence effectively with Deit-based and ViT-based backbones. The results indicate that our method can obtain more robust transferable representations than the CNN-based and ViT-based methods.

\noindent\textbf{Results on Office-31.}  Tab.~\ref{tab:office31_res} shows the quantitative comparison with the CNN-based and ViT-based methods.
Overall, our PMTrans achieves the best performance on each task with \textbf{95.3\%} accuracy and outperforms the SoTA methods with identical backbones. Numerically, PMTrans noticeably surpasses the SoTA methods with an increase of \textbf{+1.4\%} accuracy over TVT, \textbf{+2.7\%} accuracy over CDTrans, and \textbf{+1.8\%} accuracy over SSRT, respectively. 

\noindent\textbf{Results on VisDA-2017.} 
As shown in Tab.~\ref{tab:VisDA1}, our PMTrans achieves \textbf{88.0\%} accuracy and outperforms the baseline by \textbf{11.2\%}. In particular, for the `hard' categories, such as "person", our method consistently achieves a much higher performance boost from \textbf{29.0\%} to \textbf{70.3\%}. These improvements indicate that our method shows an excellent generalization capability and achieves comparable performance (\textbf{88.0\%}) with the SoTA methods (\textbf{88.7\%}). PMTrans also surpasses the SoTA methods on several sub-categories, such as "horse" and "sktbrd". 
In particular, it is shown that the SoTA methods,~\eg, CDTrans and SSRT, achieve better results on this dataset. The reason is that CDTrans and SSRT are trained with a batch size of 64 while PMTrans's batch size is 32. It indicates that when the input size is much bigger, the input can represent the data distributions better. \textit{A detailed ablation study for this issue can be found in the suppl. material.}.

\begin{figure*}[t!]
    \centering
    \includegraphics[width=0.95\textwidth]{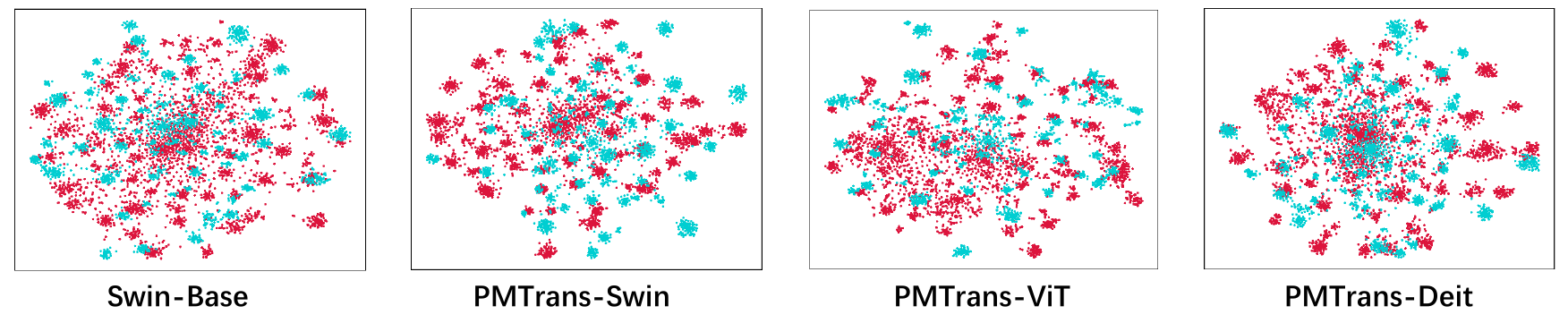}
    \vspace{-8pt}
    \caption{t-SNE visualizations for task A$\rightarrow$C on the Office-Home dataset. Source and target instances are shown in blue and red, respectively.}
    \label{fig:visual}
    \vspace{-5pt}
\end{figure*}
\begin{table*}[t]
\centering
\resizebox{0.9\linewidth}{!}{ 
\begin{tabular}{ccc|ccllllllllllllll}
\toprule
$\mathcal{L}_{cls}^{S}$&$\mathcal{L}_{f}$&$\mathcal{L}_{l}$&A$\to$ C& A$\to$ P & A $\to$ R & C $\to$ A &C $\to$ P &C $\to$ R &P$\to$ A& P$\to$ C & P$ \to$ R & R$ \to$ A &R$ \to$ C &R$ \to$ P& Avg     \\
\hline
\hline
\checkmark&&&72.7   &87.1   &90.6   &84.3   &87.3   &89.3   &80.6 &68.6 &90.3&84.8&69.4&91.3& \colorbox{lightgray}{83.6} \\
\checkmark& \checkmark&&73.9& 87.5& 91.0& 85.3 &87.9 &89.9& 82.8 &72.1& 91.2 &86.3& 74.1 &92.4&\colorbox{lightgray}{84.6}\\
\checkmark& &\checkmark&79.2&91.8&92.3&88.0&92.6&\textbf{93.0}&87.1&77.8&92.5&88.2&78.4&93.9&\colorbox{lightgray}{87.9}\\
\hline
 \checkmark & \checkmark& \checkmark&\textbf{81.3}&\textbf{92.9}&\textbf{92.8}&\textbf{88.4}&\textbf{93.4}&\textbf{93.2}&\textbf{87.9}&\textbf{80.4}&\textbf{93.0}&\textbf{89.0}&\textbf{80.9}&\textbf{94.8}&\colorbox{lightgray}{\textbf{89.0}}\\
\bottomrule
\end{tabular}}
\vspace{-5pt}
\caption{Effect of semi-supervised loss. The best performance is marked as \textbf{bold}.}
\label{tab:semi1}
\end{table*}
\begin{table*}[t!]
\centering
\resizebox{0.9\linewidth}{!}{ 
\begin{tabular}{c|clllllllllllll}
\toprule
Method  &A$\to$ C& A$\to$ P & A $\to$ R & C $\to$ A &C $\to$ P &C $\to$ R &P$\to$ A& P$\to$ C & P$ \to$ R & R$ \to$ A &R$ \to$ C &R$ \to$ P& \colorbox{lightgray}{Avg}     \\
\hline
\hline
Beta(1,1)&79.9&92.0&92.3&88.6&92.6&92.4&86.9&79.0&92.4&88.2&79.3&94.0&\colorbox{lightgray}{88.1}\\
Beta(2,2)&79.9&92.1&92.7&88.4&92.4&92.7&86.9&79.5&92.1&88.1&79.6&94.3&\colorbox{lightgray}{88.2}\\
\hline
Learning   &\textbf{81.3}&\textbf{92.9}&\textbf{92.8}&\textbf{88.4}&\textbf{93.4}&\textbf{93.2}&\textbf{87.9}&\textbf{80.4}&\textbf{93.0}&\textbf{89.0}&\textbf{80.9}&\textbf{94.8}&\colorbox{lightgray}{\textbf{89.0}}\\

\bottomrule
\end{tabular}}
\vspace{-6pt}
\caption{Effect of learning parameters. The best performance is marked as \textbf{bold}.}
\label{tab:Mixup}
\end{table*}
\vspace{-2pt}
\begin{table*}[t!]
\centering
\resizebox{0.9\linewidth}{!}{ 
\begin{tabular}{c|clllllllllllll}
\toprule
Method &A$\to$ C& A$\to$ P & A $\to$ R & C $\to$ A &C $\to$ P &C $\to$ R &P$\to$ A& P$\to$ C & P$ \to$ R & R$ \to$ A &R$ \to$ C &R$ \to$ P& \colorbox{lightgray}{Avg}     \\
\hline
\hline
Mixup &79.4&92.4&92.6&87.5&92.8&92.4&86.8&\textbf{80.3}&92.5&88.2&79.7&\textbf{95.4}&\colorbox{lightgray}{88.3}\\
CutMix&79.2&91.2&92.2&87.6&91.8&91.8&86.0&77.8&92.6&88.2&78.4&94.1&\colorbox{lightgray}{87.6}\\
\hline
PatchMix &\textbf{81.3}&\textbf{92.9}&\textbf{92.8}&\textbf{88.4}&\textbf{93.4}&\textbf{93.2}&\textbf{87.9}&\textbf{80.4}&\textbf{93.0}&\textbf{89.0}&\textbf{80.9}&94.8&\colorbox{lightgray}{\textbf{89.0}} \\
\bottomrule
\end{tabular}}
\vspace{-8pt}
\caption{Effect of PatchMix. The best performance is marked as \textbf{bold}.}
\vspace{-13pt}
\label{tab:Patch_Mixup}
\end{table*}
\noindent\textbf{Results on DomainNet.} PMTrans achieves a very high average accuracy on the most challenging DomainNet dataset, as shown in Tab. \ref{tab:domainnet}. Overall, our proposed PMTrans outperforms the SoTA methods by \textbf{+7.2\%} accuracy. Incredibly, PMTrans surpasses the SoTA methods in all the \textbf{30 sub-tasks}, which demonstrates the strong ability of PMTrans to alleviate the large domain gap. Moreover, transferring knowledge is much more difficult when the domain gap becomes significant. 
\textit{When taking more challenging $qdr$ as the target domain while others as the source domain, our PMTrans achieves an average accuracy of \textbf{27.0\%}, while ViT-based SSRT and CDTrans only achieve an average accuracy of \textbf{13.7\%} and \textbf{19.6\%}, respectively.} The comparisons on DomainNet dataset demonstrate that our PMTrans yields the best generalization ability for the challenging UDA problem.
\subsection{Ablation Study}


\noindent\textbf{Semi-supervised mixup loss.}
As shown in Tab. \ref{tab:semi1}, Swin with the semi-supervised mixup loss in the feature and label spaces outperforms the counterpart built on Swin with only source training by \textbf{+1.0\%} and \textbf{+4.3\%} on Office-Home dataset, respectively. The results indicate the effectiveness of the semi-supervised mixup loss for diminishing the domain discrepancy. Moreover, we observe that the CE loss yields better performance on the label space than that on the feature space. The reason is that the CE loss on the label space utilizing the class information performs better than on the feature space without the class information.

\noindent\textbf{Learning hyperparameters of mixup.}
Tab. \ref{tab:Mixup} shows the ablation results for the effects of learning hyperparameters of the Beta distribution on the Office-Home. We compare the learning hyperparameters of mixup with fixed parameters, such as Beta(1,1) and Beta(2,2). The proposed method achieves \textbf{+0.9\%} and \textbf{+0.8\%} accuracy increment compared with that based on Beta(1,1) and Beta(2,2). The results demonstrate that learning to estimate the distribution to build up the intermediate domain facilitates domain alignment.

\noindent\textbf{PatchMix.} Comparisons of PMTrans with Mixup \cite{ZhangCDL18} and CutMix \cite{YunHCOYC19} are shown in Tab. \ref{tab:Patch_Mixup}. PMTrans outperforms Mixup and CutMix by \textbf{+0.7\%} and \textbf{+1.4\%} accuracy on the Office-Home dataset, demonstrating that PatchMix can capture the global and local mixture information better than the global mixture Mixup and local mixture CutMix methods.

\noindent\textbf{Visualization.}
In Fig.~\ref{fig:visual}, we visualize the features learned by Swin-based, PMTrans-Swin, PMTrans-ViT, and PMTrans-Deit on task A $\rightarrow$ C from the Office-Home dataset via the t-SNE~\cite{DonahueJVHZTD14}. Compared with Swin-based and PMTrans-Swin, our PMTrans model can better align the two domains by constructing the intermediate domain to bridge them. Moreover, comparisons between PMTrans with different transformer backbones reveal that PMTrans works successfully with different backbones on UDA tasks.
\textit{Due to the page limit, more experiments and analyses can be found in the suppl. material.}
\vspace{-8pt}
\section{Conclusion and Future Work}
\vspace{-4pt}
In this paper, we proposed a novel method, PMTrans, an optimization solution for UDA from a game perspective. Specifically, we first proposed a novel ViT-based module called PatchMix that effectively built up the intermediate domain to learn discriminative domain-invariant representations for domains. And the two semi-supervised mixup losses were proposed to assist in finding the Nash Equilibria. Moreover, we leveraged attention maps from ViT to re-weight the label of each patch by its significance. PMTrans achieved the SoTA results on four benchmark UDA datasets, outperforming the SoTA methods by a large margin. In the near future, we plan to implement our PatchMix and the two semi-supervised mixup losses to solve self-supervised and semi-supervised learning problems. We will also exploit our method to tackle the challenging downstream tasks, \eg, semantic segmentation and object detection. 

\clearpage
{\small
\bibliographystyle{ieee_fullname}
\bibliography{egbib}
}

\end{document}


\title{Patch-Mix Transformer for Unsupervised Domain Adaptation: A Game Perspective

--Appendix--}


\author{Jinjing Zhu$^{1}$\thanks{These authors contributed equally to this work. }
\quad
Haotian Bai$^{1}$\footnotemark[1]
\quad
Lin Wang$^{1,2}$\thanks{Corresponding Author.}
\and
\affmark[1] AI Thrust, HKUST(GZ)\quad
\affmark[2] Dept. of CSE, HKUST\\
\quad
{\tt\small zhujinjing.hkust@gmail.com, haotianwhite@outlook.com, linwang@ust.hk}
}
\maketitle
\begin{abstract}
In this supplementary material, we first prove theorem \ref{theorem:Distribution estimation} in Section \ref{proof}. Then, Section \ref{detials} introduces the details of the proposed method, and Section \ref{algorithm} shows the algorithm of the proposed PMTrans. Section \ref{results} and Section \ref{Allation} show the results, analyses, and ablation experiments to prove the effectiveness of the proposed PMTrans. Finally, Section \ref{Discussion} shows some discussions and details about our proposed work.
\end{abstract}

\section{Proof}
\label{proof}
\subsection{Domain Distribution Estimation with PatchMix}
Let $\mathcal{H}$ denote the representation spaces with dimensionality  $\operatorname{dim}(\mathcal{H})$, $\mathcal{F}$ denote the set of encoding functions \ie, the feature extractor and $\mathcal{C}$ be the set of decoding functions \ie the classifier. Let $\mathcal{P}_{\lambda}$ be the set of functions to generate mixup ratio for building the intermediate domain. Furthermore, let $P_{S}$, $P_{T}$, and $P_{I}$ be the empirical distributions of data $\mathcal D_{s}$, $\mathcal D_{t}$, and $\mathcal D_{i}$. Define $f^{\star} \in \mathcal{F}, c^{\star} \in \mathcal{C},$ and $\mathcal{P}_{\lambda}^{\star} \in \mathcal{P}_{\lambda}$ be the minimizers of Eq. \ref{eq1} and $D(P_{S}, P_{T})$ as the measure of the domain divergence between $P_S$ and $P_T$:
\begin{equation}
\small
\begin{aligned}
\label{eq1}
 D(P_S,P_T)
 =&\inf _{f \in \mathcal{F}, c \in \mathcal{C}, \mathcal{P}_{\lambda} \in \mathcal{P}} \underset{(\boldsymbol{x}^s, \boldsymbol{y}^s),\left(\boldsymbol{x}^{t}, \boldsymbol{y}^{t}\right)} {\mathbb{E}}\\ &\ell\left(c\left(\mathcal{P}_{\lambda}\left(f(\boldsymbol{x}^{s}), f\left(\boldsymbol{x}^{t}\right)\right)\right), \mathcal{P}_{\lambda}\left(\boldsymbol{y}^{s}, \boldsymbol{y}^{t}\right)\right),   
\end{aligned}
\end{equation}
where $\ell$ is the CE loss.
Then, we can reformulate Eq.\ref{eq1} as:
\begin{align}
\small
 & D\left(P_{S},P_{T}\right)=\inf_{\boldsymbol{h}_{1}^{s}, \ldots, \boldsymbol{h}_{n}^{s} \in \mathcal{H}^{s},\boldsymbol{h}_{1}^{t}, \ldots, \boldsymbol{h}_{n}^{t} \in \mathcal{H}^{t}} \frac{1}{n_{s} \times n_{t}} \sum_{i}^{n_{s}}\sum_{j}^{n_{t}}\\
  &\left\{\inf _{c \in \mathcal{C}} \int_{0}^{1} \ell\left(f\left(\mathcal {P}_{\lambda}\left(\boldsymbol{h}_{i}^{s}, \boldsymbol{h}_{j}^{t}\right)\right), \mathcal{P}_{\lambda}\left(\boldsymbol{y}_{i}^{s}, \boldsymbol{y}_{j}^{t}\right)\right) p(\lambda) \mathrm{d} \lambda\right\},   \nonumber
\end{align}
where $\boldsymbol{h}_{i}^{s} = f(\boldsymbol{x}^{s}_{i})$ and $\boldsymbol{h}_{j}^{t} = f(\boldsymbol{x}^{t}_{j})$.
Inspired and borrowed by this work\cite{VermaLBNMLB19}, we give proof as follows.

\begin{theorem}

\label{theorem:Distribution estimation}:Let $d \in \mathbb{N}$ to represent the number of classes contained in three sets $\mathcal D_{s}$, $\mathcal D_{t}$, and $\mathcal D_{i}$. If $\operatorname{dim}(\mathcal{H}) \geq d-1$, ${\mathcal{P}_{\lambda}}’ \ell(c^{\star}(f^{\star}(\boldsymbol{x}_{i})),\boldsymbol{y}^{s})+ (1-{\mathcal{P}_{\lambda}}')\ell(c^{\star}(f^{\star}(\boldsymbol{x}_{i})),\boldsymbol{y}^{t})=0$, then $D\left(P_{S},P_{T}\right)=0$ and the corresponding minimizer $c^{\star}$ is a linear function from $\mathcal{H}$ to $\mathbb{R}^{d}$. Denote the scaled mixup ratio sampled from a learnable Beta distribution as $\mathcal{P}_{\lambda}’$.

\end{theorem}

\noindent\textbf{Proof}:
First, the following statement is $\operatorname{true}$ if $\operatorname{dim}(\mathcal{H}) \geq d-1$ :
$$
\exists A, H \in \mathbb{R}^{\operatorname{dim}(\mathcal{H}) \times d}, b \in \mathbb{R}^{d}: A^{\top} H+b _{d}^{\top}=I_{d \times d},
$$
where $I_{d \times d}$ and $1_{d}$ denote the $d$-dimensional identity matrix and all-one vector, respectively. In fact, $b _{d}^{\top}$ is a rank-one matrix, and the rank of the identity matrix is $d$. Since the column set span of $I_{d \times d}$ needs to be contained in the span of $A^{\top}H+b _{d}^{\top}$, $A^{\top}H$ only needs to be a matrix with the rank $d-1$ to meet this requirement.

Let $c^{\star}(h)=A^{\top} h+b$, for all $h \in \mathcal{H}$.  Let $f^{\star}\left(\boldsymbol{x}_{i}^{s}\right)=H_{\zeta_{i}^{s},:}$ and $f^{\star}\left(\boldsymbol{x}_{j}^{t}\right)=H_{\zeta_{j}^{t},:}$ be the $\zeta_{i}$-th and $\zeta_{j}$-th slice of $H$, respectively.  
Specifically, $\zeta_{i}^{s}, \zeta_{i}^{t} \in\{1, \ldots, d\}$ stand for the class-index of the examples $\boldsymbol{x}_{i}^{s}$ and $\boldsymbol{x}_{j}^{t}$. 
Given Eq.\ref{eq1}, the intermediate domain sample $\boldsymbol{x}^{i}_{ij} = \mathcal {P}_{{\lambda}}^{\star}(\boldsymbol{a}_i, \boldsymbol{b}_j) = \mathcal {P}_{{\lambda}}'\cdot \boldsymbol{a}_i + \left(1-{P}_{{\lambda}}'\right)\cdot \boldsymbol{b}_j$, and the definition of cross-entropy loss $\ell$, we get:
\begin{equation}
    \label{eq:loss_merge}
    \begin{aligned}
    &{{\mathcal {P}_{{\lambda}}}'} \ell\left(c^{\star}\left(f^{\star}\left(\boldsymbol{x}^{i}_{ij}\right)\right), \boldsymbol{y}^{s}_{i}\right)+(1-{\mathcal{P}_{\lambda}}')\ell\left(c^{\star}\left(f^{\star}\left(\boldsymbol{x}^{i}_{ij}\right)\right), \boldsymbol{y}^{t}_{j}\right)
    \\
    &=\ell\left(c^{\star}\left({\mathcal {P}_{{\lambda}}}^{\star}\left(f^{\star}\left(\boldsymbol{x}_{i}^{s}\right), f^{\star}\left(\boldsymbol{x}_{j}^{t}\right)\right)\right), {\mathcal {P}_{{\lambda}}}^{\star}\left(\boldsymbol{y}_{i}^{s}, \boldsymbol{y}_{j}^{t}\right)\right)\\
    &= D\left(P_{S},P_{T}\right)=0.
    \end{aligned}
\end{equation} 
%
%
%
 %
Eq.~\ref{eq:loss_merge} reveals that the source and target domains are aligned if mixing the patches from two domains is equivalent to mixing the corresponding labels. 

Furthermore, we see the following: 
\begin{figure*}[t]
     \centering
     \begin{subfigure}[b]{0.4\textwidth}
         \centering
         \includegraphics[width=\textwidth]{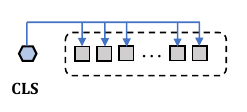}
         (a)\\
    \label{fig:has_cls}
     \end{subfigure}
     \begin{subfigure}[b]{0.4\textwidth}
         \centering
         \includegraphics[width=\textwidth]{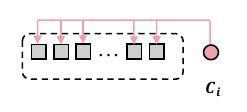}
         (b)\\
    \label{fig:no_cls}
     \end{subfigure}
        \caption{(a) Deit/ViT attention scores with the CLS token. (b) Swin attention scores with an output unit of Classifier that refers to $C_i$ . The dashed line denotes the sequence with each square representing a patch.}
\label{fig:attention}
\end{figure*}
\begin{equation}
    \begin{aligned}
  &\ell\left(c^{\star}\left({\mathcal {P}_{\lambda}}^{\star}\left(f^{\star}\left(\boldsymbol{x}_{i}^{s}\right), f^{\star}\left(\boldsymbol{x}_{j}^{t}\right)\right)\right), {\mathcal {P}_{\lambda}}^{\star}\left(\boldsymbol{y}_{i}^{s}, \boldsymbol{y}_{j}^{t}\right)\right)\\
  &=  \ell\left(A^{\top} {\mathcal {P}_{\lambda}}^{\star}\left(H^{s}_{\zeta_{i},:}, H^{t}_{\zeta_{j},:}\right)+b, {\mathcal {P}_{\lambda}}^{\star}\left(\boldsymbol{y}^{s}_{i, \zeta_{i}}, \boldsymbol{y}^{t}_{j, \zeta_{j}}\right)\right)\\
  &=0.  
    \end{aligned}
\end{equation}
The result follows from $A^{\top} H_{\zeta_{i}^{s},:}+b=y_{i, \zeta_{i}^{s}} \text { for all } i \text {, }$ and $A^{\top} H_{\zeta_{j}^{t},:}+b=y_{i, \zeta_{j}^{t}} \text { for all } j \text {. }$ 
If $\operatorname{dim}(\mathcal{H}) \geq d-1$, $f^{\star}\left(\boldsymbol{x}_{i}^{t}\right)$ and $f^{\star}\left(\boldsymbol{x}_{j}^{t}\right)$ in the representation space $\mathcal{H}$ have some degrees of freedom to move independently. It also implies that when Eq.~\ref{eq:loss_merge} is minimized, the representation of each class lies on a subspace of dimension $\operatorname{dim}(\mathcal{H})-d+1$, and with larger $dim(\mathcal{H})$, the majority of directions in $\mathcal{H}$-space will contain zero variance in the class-conditional manifold.

In practice, we utilize Eq.~\ref{eq:loss_merge} to measure the domain gaps between the intermediate domain, and other domains and decrease them in the label and feature space, as illustrated in the main paper.
%

\section{Details}
\label{detials}
\subsection{Datasets}
To evaluate the proposed method, we conduct extensive experiments on four popular UDA benchmarks, including  Office-31 \cite{SaenkoKFD10}, Office-Home \cite{VenkateswaraECP17}, VisDA-2017 \cite{abs-1710-06924}, and DomainNet \cite{PengBXHSW19}. 

\textbf{Office-31} consists of 4110 images of 31 categories, with three domains: Amazon (A), Webcam (W), and DSLR (D).

\textbf{Office-Home} is collected from four domains: Artistic images (A), Clip Art (C), Product images (P), and Real-World images (R) and consists of 15500 images from 65 classes. 

\textbf{VisDA-2017} is a more challenging dataset for synthetic-to-real domain adaptation. We set 152397 synthetic images as the source domain data and 55388 real-world images as the target domain data. 

\textbf{DomainNet} is a large-scale benchmark dataset, which has 345 classes from six domains (Clipart (clp), Infograph (inf), Painting (pnt), Quickdraw (qdr), Real (rel), and Sketch (skt)).
\subsection{Attention Map}
\label{sec:attention map}
We calculate the attention score in two ways based on whether the CLS token is present in the sequence. For Swin Transformer, we adopt a method similar to CAM \cite{DBLP:journals/corr/ZhouKLOT15} instead of changing the backbone from CNN to Transformer. Specifically, for a given image, let $f_k(x,y)$ represent the encoded patch $k$ in the last layer at spatial location $(x,y)$. The output of Transformer is followed by a global average pooling (GAP) layer$\sum(x, y)$ and a linear classification head. For the specific class $C_i$, the classification score $S_{C_i}$ is:
\begin{equation}
\label{eq:cls_score}
    S_{C_i} = \sum_{j}w_j^{C_i}\sum_{x,y}f_k(x,y),
\end{equation}
where $w_j^{C_i}$ represents the weight corresponding to class $C_i$ for unit $j$ in the hidden dimension. Eq.\ref{eq:cls_score} ensembles the semantics over both spatial contexts $\sum(x, y)$ and the linear head units $\sum(j)$. Then given Eq.\ref{eq:cls_score}, as shown in Fig.\ref{fig:attention} \textcolor{red}{(b)}, for a given $C_i$, we reallocate the semantic information from the output of linear head unit of $C_i$. In detail, we define the semantic activation map at location $(x,y)$ for a specific class $C_i$ as:

\[
    M_{C_i}(x,y) = \sum_j w_j^{C_i} f_k(x,y),
\]
where $M_{C_i}\in \mathbb{R}^2$ is the activation for class $C_i$, and we infer $C_i$ by the ground-truth label in the source domain and the pseudo-label in the target domain to obtain the corresponding class activation map to build the intermediate domain. Then, we use $M_{C_i}$ as the attention map after the softmax operation. 

On the other hand, when the CLS token is present in the output sequence of Transformer like Deit/ViT, we simply take the attention scores from the self-attention, \textit{i.e.} the similarity matrix of each layer $i$ in Transformer $Attn_i \in \mathcal{R}^{H\times N\times N}$, and take the average in the head dimension $H$:
\[
    Attn_i = \frac{1}{H}\sum_h Attn_h,
\]
where $N$ is the sequence length. Next, we only take the CLS token's attention after the softmax operation, as shown in  Fig.\ref{fig:attention} \textcolor{red}{(a)}, and then summarize each layer's scores to obtain the final attention scores $Attn$. 
\[
    Attn = \frac{1}{I}\sum_i Attn_i.
\]
\begin{figure}[t]
    \centering
    \includegraphics[width=\linewidth]{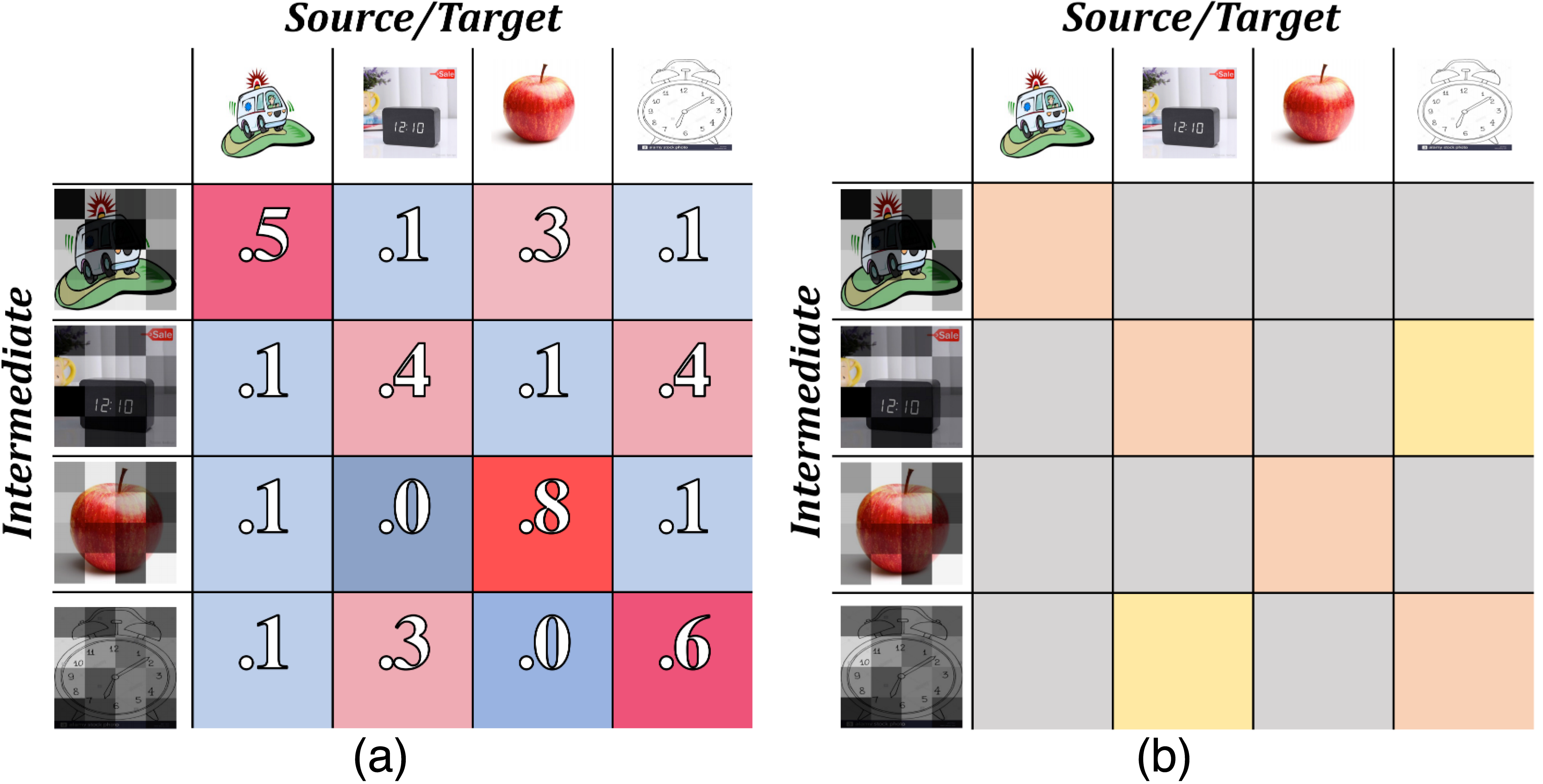}
    \caption{Illustration of the semi-supervised loss in feature space.}
    \label{fig:align_feature}
\end{figure}
\subsection{Semi-supervised mixup loss in the feature space}
\label{detial of semi-supervised loss}
In Fig.\ref{fig:align_feature}, we illustrate the semi-supervised loss in the feature space by similarity between features (in Fig. \ref{fig:align_feature} {\color{red}(a)}) and label spaces(in Fig.\ref{fig:align_feature} {\color{red}(b)}). To compute the similarity of features,  we use the normalized cosine similarity loss between the intermediate domain (column) and source/target domain(row) in the feature space, as shown in Fig.\ref{fig:align_feature}{\color{red}(a)}. Each row denotes the normalized similarity between a sample of the intermediate domain and counterparts from the source domains. For example, we first use the cosine similarity to calculate the similarities between one intermediate sample "car" and four sources (or target) samples (car, clock, apple, sketch clock). Then we normalize these similarities. As for the similarity of outputs (or) labels, since the source samples are labeled, and the target samples are unlabeled, we design two different methods to calculate the supervised and unsupervised label similarities. As for the label similarity between the intermediate and source domains, the intermediate and source samples both share the same labels. Therefore, we define the label similarity $\boldsymbol{y}^{is} = \boldsymbol{y}^s({\boldsymbol{y}^s})^{\intercal}$, as shown in Fig.\ref{fig:align_feature} {\color{red}(b)}. Specifically, $\boldsymbol{y}^{is}$, denoted by the yellow and pink colors, indicates that the label similarity between samples is one for these samples with the same labels (zero for different labels). For example, the label similarities between one intermediate sample "sketch clock" and four sources (or target) samples car, clock, apple, and sketch clock are zero, one, zero, and one. As for the label similarity between the intermediate and target samples, we only know that the intermediate and source samples both share overlapped patches due to lack of supervision. Therefore, the label similarity $\boldsymbol{y}^{it}$ between samples with overlapped patches should be one (pink color), and others should be zero. And we define the label similarity $\boldsymbol{y}^{it}$ as identity matrix. For example, the label similarities between one unlabeled intermediate sample "sketch clock" and four unlabeled target samples car, real clock, apple, and sketch clock are zero, zero, zero, and one. After obtaining the feature and label similarities, we utilize the CE loss $\ell$ to measure the discrepancy between these similarities as the domain gap between the intermediate and other domains.

\subsection{Optimization}
In our game, $m$-th player is endowed with a cost function $J_{m}$ and strives to reduce its cost, which contributes to the change of CE. We now define each player's cost function $J_{m}$ as
\begin{equation}
\small
\begin{split}\label{eq:objective}
J_{{\mathcal{F}}}\left(\boldsymbol{\omega}_{\mathcal{F}}, \boldsymbol{\omega}_{-{\mathcal{F}}}\right) &:=\mathcal{L}_{cls}^{S}(\boldsymbol{\omega}_{\mathcal{F}}, \boldsymbol{\omega}_{\mathcal{C}})+\alpha {CE}_{s,i, t}(\boldsymbol{\omega}),\\ 
J_{{\mathcal{C}}}\left(\boldsymbol{\omega}_{\mathcal{C}}, \boldsymbol{\omega}_{-{\mathcal{C}}}\right) &:=\mathcal{L}_{cls}^{S}(\boldsymbol{\omega}_{\mathcal{F}}, \boldsymbol{\omega}_{\mathcal{C}})+\alpha {CE}_{s,i, t}(\boldsymbol{\omega}),\\ 
J_{{\mathcal{P}}}\left(\boldsymbol{\omega}_{{\mathcal{P}}}, \boldsymbol{\omega}_{-{\mathcal{P}}}\right) &:=-\alpha {CE}_{s,i, t}(\boldsymbol{\omega}),    
\end{split}
\end{equation}
where $\alpha$ is the trade-off parameter, $\ell$ is the supervised classification loss for the source domain, and $\text{CE}_{s,i,t}(\boldsymbol{\omega})$ is the discrepancy between the intermediate domain and the source/target domain.
To clarify the min-max process, we introduce the game's vector field $v(w)$, which is identical to the gradient for every player.
\begin{definition}
\label{Def:vector field}
(Vector field): .
\[
v(\boldsymbol{\omega}) := (\bigtriangledown_{{\boldsymbol{\omega}}_\mathcal{F}}J_\mathcal{F}, \bigtriangledown_{{\boldsymbol{\omega}}_\mathcal{C}}J_\mathcal{C}, \bigtriangledown_{{\boldsymbol{\omega}}_\mathcal{P}}J_\mathcal{P})
\]
\end{definition}
By examining Definition.\ref{Def:vector field} with respect to Eq.(\ref{eq:objective}), the process can be categorized into both cooperation and competition \cite{abs-2202-05352}. 

\begin{equation}
    v(w)=
    \begin{pmatrix}
    \bigtriangledown_{{\boldsymbol{\omega}}_\mathcal{F}}\mathcal{L}_{cls}^{S}(\boldsymbol{\omega}_\mathcal{F}, \boldsymbol{\omega}_\mathcal{C})\\
    \bigtriangledown_{{\boldsymbol{\omega}}_\mathcal{C}}\mathcal{L}_{cls}^{S}(\boldsymbol{\omega}_{\mathcal{F}}, \boldsymbol{\omega}_{\mathcal{C}})\\
    0
    \end{pmatrix}
    +
    \begin{pmatrix}
    \alpha\bigtriangledown_{{\boldsymbol{\omega}}_\mathcal{F}}\text{CE}_{s,i, t}(\boldsymbol{\omega})\\
    \alpha\bigtriangledown_{{\boldsymbol{\omega}}_\mathcal{C}}\text{CE}_{s,i, t}(\boldsymbol{\omega})\\
    -\alpha \bigtriangledown_{{\boldsymbol{\omega}}_\mathcal{P}}\text{CE}_{s,i, t}(\boldsymbol{\omega})
    \end{pmatrix},   
\end{equation}
where the left part is related to the gradient of $\mathcal{L}_{cls}^{S}(\boldsymbol{\omega}_{\mathcal{F}}, \boldsymbol{\omega}_{\mathcal{C}})$, and the right part denotes the adversarial behavior on producing or consuming CE in the network. In this Min-max CE Game, each player behaves selfishly to reduce their cost function. This competition on the network's CE will possibly end with a situation where no one has anything to gain by changing only one's strategy, called NE. Note that our method does not require explicit usage of gradient reverse layers as the prior GAN-based game design~\cite{GaninL15}. Our training is optimized as 

\begin{equation}
\label{optimal_one}
v(\boldsymbol{\omega})=\bigtriangledown_{(\boldsymbol{\omega}_{\mathcal{F}},\boldsymbol{\omega}_{\mathcal{C}})}\mathcal{L}_{cls}^{S}(\boldsymbol{\omega}_{\mathcal{F}}, \boldsymbol{\omega}_{\mathcal{C}})+\alpha\bigtriangledown_{{\boldsymbol{\omega}}}{CE}_{s, i,t}(\boldsymbol{\omega}).
\end{equation}
\subsection{Comparisons with Mixup variants}
\begin{figure*}[t]
    \centering
    \includegraphics[width=0.99\linewidth]{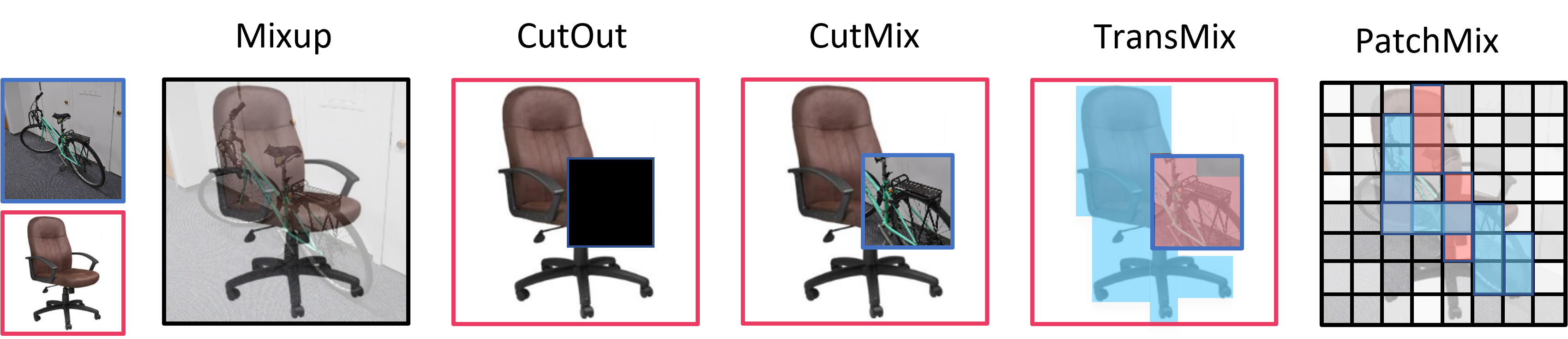}
    \caption{PMTrans and Mixup variants}
    \label{fig:PMTrans_mixup}
\end{figure*}

In Fig.~\ref{fig:PMTrans_mixup}, we show the visual comparisons between the PatchMix and mainstream Mixup variants. Mixup \cite{ZhangCDL18} mixes two samples by interpolating both the images and labels, which suffers from the local ambiguity. CutOut \cite{abs-1708-04552} proposes to randomly mask out square regions of input during training to improve the robustness of the CNNs. Since CutOut decreases the ImageNet localization or object detection performances, CutMix \cite{YunHCOYC19} is further introduced to randomly cut and paste the regions in an image, where the ground truth labels are also mixed proportionally to the area of the regions. However, sometimes there is no valid object in the mixed image due to the random process in augmentation, but there is still a response in the label space. Therefore, not all pixels are created equal, and the labels of pixels should be re-weighted. TransMix \cite{DBLP:journals/corr/abs-2111-09833} is proposed to utilize the attention map to assign the confidence for the mixed samples and re-weighted the labels of pixels. In comparison, we unify these global and local mixup techniques in our PatchMix by learning to combine two patches to form a mixed patch and obtain mixed samples. Furthermore, we also learn the hyperparameters of the mixup ratio for each patch and effectively build up the intermediate domain samples.

\section{Algorithm}
\label{algorithm}

In summary, the whole algorithm to train the proposed PMTrans is shown in Algorithm \ref{algo:algo}.

\begin{algorithm}[htb] 
\caption{Patch-Mix Transformer for Unsupervised Domain
Adaptation} 
\label{algo:algo} 
\begin{algorithmic}[1] 
\REQUIRE source domain data $\mathcal{D}_{s}$ and target domain data $\mathcal{D}_{t}$. \\
\ENSURE learned parameters of feature extractor $\mathcal{\boldsymbol{\omega}}_{\mathcal{F}}$, classifier $\mathcal{\boldsymbol{\omega}}_{\mathcal{C}}$, and PatchMix $\mathcal{\boldsymbol{\omega}}_{\mathcal{P}}$. \\
\FOR{$k=0$ to $\mathrm{MaxIter}$}
\STATE Sample a batch of input from source data and target data.
\STATE Encode the patches of source and target inputs by the patch embedding (Emb) layer.
\STATE Calculate the normalized attention score for each patch as Section \ref{sec:attention map}.
\STATE Sample the mixup ratio from Beta($\mathcal{\boldsymbol{\omega}}_{\mathcal{P}}$)
\STATE Construct the intermediate domain input as shown in Eq.~\ref{eq:mixup_eq}.
\STATE Calculate the semi-supervised mixup loss in the feature space via Eq.~\ref{eq:mixupinfeature}.
\STATE Calculate the semi-supervised mixup loss in the label space via Eq. \ref{eq:mixupinlabel}.
\STATE Measure the domain divergence between intermediate domain and other two domains via Eq.~\ref{eq:total }. 
\STATE Update network parameters $\boldsymbol{\omega}$ by optimization (\ref{optimal_one}) via a AdamW \cite{LoshchilovH19} optimizer.
\ENDFOR
\RETURN $\mathcal{\boldsymbol{\omega}}_{\mathcal{F}}$, $\mathcal{\boldsymbol{\omega}}_{\mathcal{C}}$, and $\mathcal{\boldsymbol{\omega}}_{\mathcal{P}}$
\end{algorithmic}
\end{algorithm}
where the related loss functions are shown as follows.
\begin{equation}
\label{eq:mixup_eq}
\small
\begin{split}
  \boldsymbol{x}^i= \mathcal{P}_\lambda(\boldsymbol{x}^s, \boldsymbol{x}^t), &~\boldsymbol{x}^{i}_{k}=\lambda_{k} \odot \boldsymbol{x}_k^s + (1-\lambda_k)\odot\boldsymbol{x}_k^t, \\
\boldsymbol{y}^i=\mathcal{P}_\lambda(\boldsymbol{y}^s, \boldsymbol{y}^t) &= \lambda^s\boldsymbol{y}^s + \lambda^t\boldsymbol{y}^t.
\end{split}
\end{equation}

\begin{equation}
\small
\label{eq:lambda}
\begin{split}
 \lambda^s &= \frac{\sum_{k=1}^{n} \lambda_{k} {a}_{k}^{s}}{\sum_{k=1}^{n} \lambda_{k} {a}_{k}^{s}+\sum_{k=1}^{n}(1-\lambda_{k}){a}_{k}^{t}}
, \\
\lambda^t &= \frac{\sum_{k=1}^{n}(1-\lambda_{k}) {a}_{k}^{t}}{\sum_{k=1}^{n} \lambda_{k} {a}_{k}^{s}+\sum_{k=1}^{n}(1-\lambda_{k}){a}_{k}^{t}}.
\end{split}
\end{equation}
\begin{equation}
\label{eq:mixupinfeature}
\small
\begin{split}
    \mathcal{L}_{f}^{I,S}(\boldsymbol{\omega}_{\mathcal{F}}, \boldsymbol{\omega}_{\mathcal{P}}) = \mathbb{E}_{\left(\boldsymbol{x}^{i}, \boldsymbol{y}^{i}\right) \sim D^{i}} \lambda^{s}\ell\left(d(\boldsymbol{x}^{i},\boldsymbol{x}^{s}), \boldsymbol{y}^{is}\right), \\
 \mathcal{L}_{f}^{I,T}(\boldsymbol{\omega}_{\mathcal{F}}, \boldsymbol{\omega}_{\mathcal{P}}) = \mathbb{E}_{\left(\boldsymbol{x}^{i}, \boldsymbol{y}^{i}\right) \sim D^{i}} \lambda^{t}\ell\left(d(\boldsymbol{x}^{i}, \boldsymbol{x}^{t}), \boldsymbol{y}^{it}\right).   
\end{split}
\end{equation}
\begin{equation}
\label{eq:mixupinlabel}
\small
\begin{split}
\mathcal{L}_{l}^{I,S}(\boldsymbol{\omega}) &= \mathbb{E}_{\left(\boldsymbol{x}^{i}, \boldsymbol{y}^{i}\right) \sim D^{i}} \lambda^{s} \ell\left(\mathcal{C}\left(\mathcal {F}\left(\boldsymbol{x}^{i}\right)\right), \boldsymbol{y}^{s}\right),\\
\mathcal{L}_{l}^{I,T}(\boldsymbol{\omega}) &= \mathbb{E}_{\left(\boldsymbol{x}^{i}, \boldsymbol{y}^{i}\right) \sim D^{i}} \lambda^{t} \ell\left(\mathcal{C}\left(\mathcal {F}\left(\boldsymbol{x}^{i}\right)\right), \hat{\boldsymbol{y}^{t}}\right).
\end{split}
\end{equation}  
\begin{equation}
\label{eq:CE}
     {\text{CE}}_{s,i,t}(\boldsymbol{\omega}) =  \mathcal{L}_{f}(\boldsymbol{\omega}_{\mathcal{F}}, \boldsymbol{\omega}_{\mathcal{P}})+\mathcal{L}_{l}(\boldsymbol{\omega}).
\end{equation}
\begin{equation}
\label{eq:total }
 J\left(\boldsymbol{\omega}\right) :=\mathcal{L}_{cls}^{S}(\boldsymbol{\omega}_{\mathcal{F}}, \boldsymbol{\omega}_{\mathcal{C}})+\alpha {\text{CE}}_{s,i,t}(\boldsymbol{\omega}).
\end{equation}
Note that these above equations are introduced in detail in the main paper. 

\section{Results and Analyses}
\label{results}
\subsection{The comparisons on the Office-31, Office-Home, and VisDA-2017}

We compare PMTrans with the SoTA methods, including ResNet- and ViT-based methods. The ResNet-based methods are FixBi \cite{NaJCH21}, CGDM \cite{Du0S0021}, MCD \cite{SaitoWUH18}, SWD \cite{LeeBBU19}, SCDA \cite{0008XLLLQL21}, BNM \cite{CuiWZLH020}, MDD \cite{0002LLJ19}, CKB \cite{LuoR21}, TSA \cite{LiXGLWL21}, DWL \cite{XiaoZ21}, ILA \cite{SharmaKC21}, Symnets \cite{ZhangTJT19}, CAN \cite{Kang0YH19}, and PCT  \cite{TanwisuthFZZZCZ21}. The ViT-based methods are SSRT \cite{abs-2204-07683}, CDTrans \cite{abs-2109-06165}, and TVT \cite{abs-2108-05988}, and we directly quote the results in their original papers for fair comparison. And the more detailed comparisons on three datasets are shown as Tab. \ref{tab:office31_res}, Tab. \ref{tab:officeHome_Res}
, and Tab. \ref{tab:VisDA_result}.
\begin{figure}
    \centering
    \includegraphics[width=0.8\linewidth]{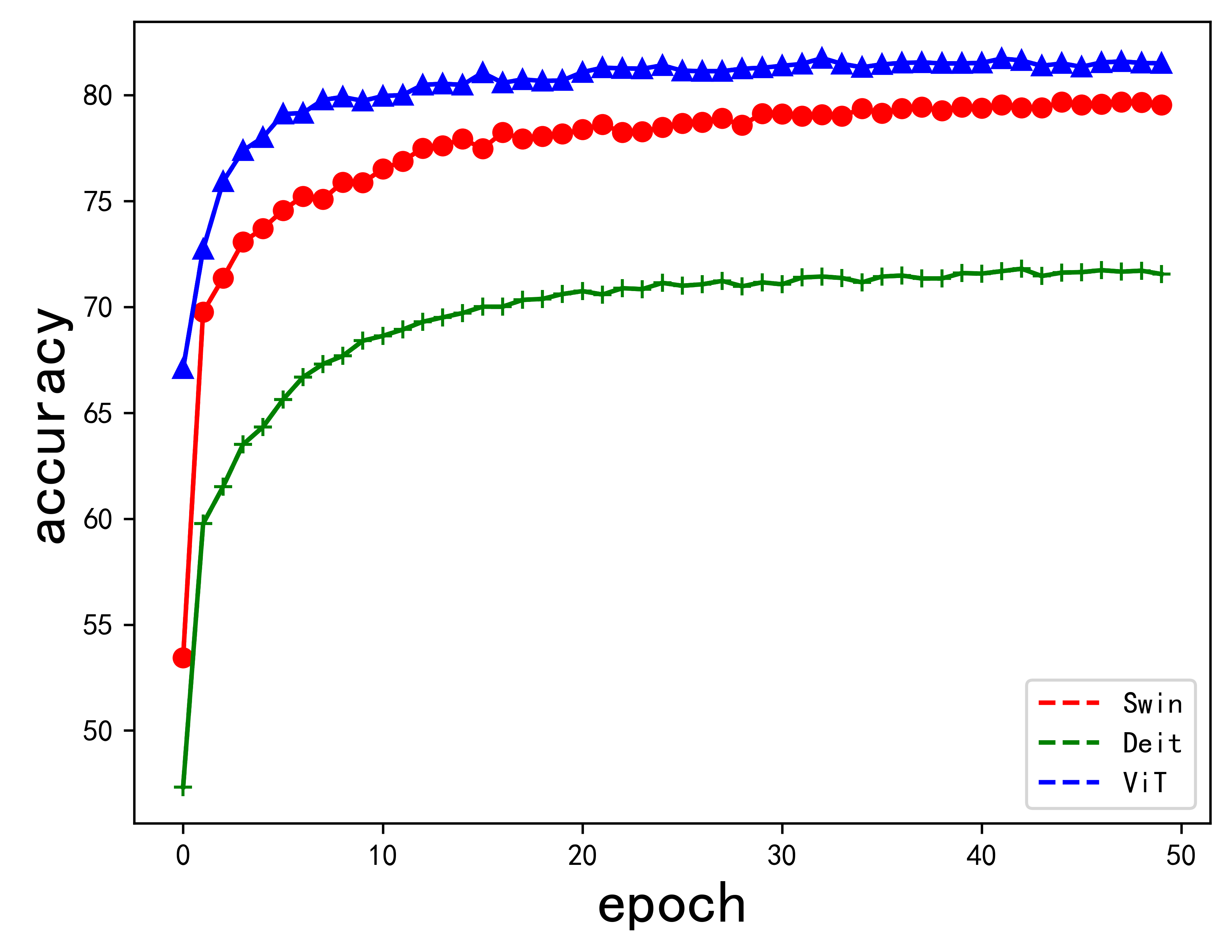}
    \caption{Accuracy on the task A $\to$ C (Office-Home)}
    \label{fig:my_label}
\end{figure}

\begin{table}[t]
\centering
\resizebox{\linewidth}{!}{
\begin{tabular}{c|l|lllllll}
\toprule
Method  &&A $\to$ W & D$\to$ W & W$\to$ D & A$\to$ D &D$\to$ A &W$\to$ A & \colorbox{lightgray}{Avg}  \\
\hline
\hline
ResNet-50 &\multirow{9}{*}{\rotatebox{90}{ResNet}} &  68.9& 68.4 &62.5& 96.7& 60.7& 99.3& \colorbox{lightgray}{76.1}   \\
BNM &&91.5&98.5&\textbf{100.0}&90.3&70.9&71.6&\colorbox{lightgray}{87.1}\\
DWL& &89.2 &99.2& \textbf{100.0}& 91.2& 73.1& 69.8& \colorbox{lightgray}{87.1}\\
MDD&&94.5& 98.4& \textbf{100.0}& 93.5& 74.6 &72.2& \colorbox{lightgray}{88.9}\\
TSA&&94.8& 99.1& \textbf{100.0}& 92.6& 74.9& 74.4& \colorbox{lightgray}{89.3}\\
ILA+CDAN&&95.7& 99.2& \textbf{100.0} &93.4 &72.1 &75.4& \colorbox{lightgray}{89.3}\\
PCT&&94.6& 98.7& 99.9& 93.8& 77.2& 76.0& \colorbox{lightgray}{90.0}\\
SCDA&&94.2 &98.7& 99.8& 95.2& 75.7& 76.2& \colorbox{lightgray}{90.0}\\
FixBi &  &  96.1& 99.3& \textbf{100.0}& 95.0&78.7& 79.4& \colorbox{lightgray}{91.4}  \\ 
\midrule
TVT &\multirow{7}{*}{\rotatebox{90}{ViT}}&96.4& 99.4& \textbf{100.0}& 96.4& 84.9& 86.0&\colorbox{lightgray}{93.9}   \\
Deit-Base &     &89.2& 98.9& \textbf{100.0}& 88.7& 80.1& 79.8& \colorbox{lightgray}{89.5}\\
CDTrans-Deit& & 96.7 &99.0& \textbf{100.0}& 97.0& 81.1& 81.9 &\colorbox{lightgray}{92.6}  \\
\textbf{PMTrans-Deit}&&99.0&99.4&\textbf{100.0}&96.5&81.4&82.1&\colorbox{lightgray}{93.1}\\
ViT-Base       &&91.2& 99.2& \textbf{100.0}& 90.4& 81.1& 80.6& \colorbox{lightgray}{91.1}  \\
SSRT-ViT&&97.7& 99.2 &\textbf{100.0} &98.6& 83.5& 82.2& \colorbox{lightgray}{93.5}\\
\textbf{PMTrans-ViT}&& 99.1&\textbf{99.6}&\textbf{100.0}&99.4&85.7&86.3&\colorbox{lightgray}{95.0}\\
\midrule
Swin-Base &\multirow{2}{*}{\rotatebox{90}{Swin}}&97.0   &99.2   &\textbf{100.0}   &95.8   &82.4   &81.8   &\colorbox{lightgray}{92.7}   \\
\textbf{PMTrans-Swin} & &\textbf{99.5}   &99.4  &\textbf{100.0}   &\textbf{99.8}   &\textbf{86.7}   &\textbf{86.5}   &\colorbox{lightgray}{\textbf{95.3}}  \\ 
\bottomrule
\end{tabular}}
\caption{Comparison with SoTA methods on Office-31. The best performance is marked as \textbf{bold}.}
\label{tab:office31_res}
\vspace{-10pt}
\end{table}
\begin{table*}[t]
\centering
\resizebox{0.99\linewidth}{!}{ 
\begin{tabular}{c|l|llllllllllllll}
\toprule
Method &&A$\to$ C& A$\to$ P & A $\to$ R & C $\to$ A &C $\to$ P &C $\to$ R &P$\to$ A& P$\to$ C & P$ \to$ R & R$ \to$ A &R$ \to$ C &R$ \to$ P& Avg     \\
\hline
\hline
ResNet-50 &\multirow{9}{*}{\rotatebox{90}{ResNet}}& 44.9& 66.3 &74.3 &51.8& 61.9 &63.6& 52.4 &39.1& 71.2& 63.8& 45.9& 77.2 &\colorbox{lightgray}{59.4}   \\
MCD &&48.9& 68.3& 74.6& 61.3& 67.6& 68.8& 57.0& 47.1 &75.1& 69.1& 52.2& 79.6& \colorbox{lightgray}{64.1}\\
Symnets&&47.7& 72.9 &78.5& 64.2 &71.3& 74.2& 64.2& 48.8& 79.5& 74.5& 52.6 &82.7& \colorbox{lightgray}{67.6}\\
MDD &&54.9& 73.7& 77.8& 60.0& 71.4& 71.8& 61.2& 53.6& 78.1& 72.5& 60.2& 82.3& \colorbox{lightgray}{68.1}\\
TSA&&53.6& 75.1& 78.3& 64.4 &73.7& 72.5& 62.3& 49.4 &77.5& 72.2& 58.8 &82.1& \colorbox{lightgray}{68.3}\\
CKB&&54.7 &74.4 &77.1& 63.7& 72.2& 71.8 &64.1& 51.7 &78.4 &73.1& 58.0& 82.4& \colorbox{lightgray}{68.5}\\
BNM&&56.7 &77.5& 81.0& 67.3& 76.3& 77.1& 65.3& 55.1& 82.0 &73.6 &57.0& 84.3& \colorbox{lightgray}{71.1}\\
PCT&& 57.1 & 78.3&81.4 & 67.6&77.0&76.5& 68.0& 55.0&81.3 &74.7& 60.0 & 85.3& \colorbox{lightgray}{71.8}\\
FixBi && 58.1& 77.3& 80.4 &67.7& 79.5& 78.1& 65.8& 57.9& 81.7& 76.4& 62.9& 86.7& \colorbox{lightgray}{72.7}  \\
\midrule
TVT & \multirow{3}{*}{\rotatebox{90}{ViT}}&   74.9&86.8&89.5&82.8&88.0&88.3&79.8&71.9&90.1&85.5&74.6&90.6&\colorbox{lightgray}{83.6}\\
Deit-Base&&61.8 &79.5& 84.3& 75.4 &78.8& 81.2& 72.8 &55.7& 84.4& 78.3& 59.3& 86.0 &\colorbox{lightgray}{74.8} \\
CDTrans-Deit&  & 68.8&85.0&86.9&81.5&87.1&87.3&79.6&63.3&88.2&82.0&66.0&90.6&\colorbox{lightgray}{80.5}   \\
\textbf{PMTrans-Deit} &&71.8& 87.3 &88.3 &83.0 &87.7 & 87.8 &78.5 &67.4 &89.3 &81.7 &70.7 & 92.0 &\colorbox{lightgray}{82.1}\\
ViT-Base &&67.0 &85.7 &88.1 &80.1 &84.1 &86.7 &79.5 &67.0 &89.4 &83.6 &70.2 &91.2 & \colorbox{lightgray}{81.1}\\
SSRT-ViT&&75.2& 89.0& 91.1& 85.1 &88.3& 89.9& 85.0& 74.2 &91.2& 85.7& 78.6 &91.8& \colorbox{lightgray}{85.4}\\
\textbf{PMTrans-ViT} &&81.2&91.6&92.4&\textbf{88.9}&91.6&93.0&\textbf{88.5}&80.0&\textbf{93.4}&\textbf{89.5}&\textbf{82.4}&94.5&\colorbox{lightgray}{88.9}\\
\midrule
Swin-Base&\multirow{2}{*}{\rotatebox{90}{Swin}}&72.7   &87.1   &90.6   &84.3   &87.3   &89.3   &80.6 &68.6 &90.3&84.8&69.4&91.3& \colorbox{lightgray}{83.6}  \\
\textbf{PMTrans-Swin} && \textbf{81.3}&\textbf{92.9}&\textbf{92.8}&88.4&\textbf{93.4}&\textbf{93.2}&87.9&\textbf{80.4}&93.0&89.0&80.9&\textbf{94.8}&\colorbox{lightgray}{\textbf{89.0}}\\
\bottomrule
\end{tabular}}
\caption{Comparison with SoTA methods on Office-Home. The best performance is marked as \textbf{bold}.}
\label{tab:officeHome_Res}
\end{table*}
\begin{table*}[t]
\centering
\resizebox{\linewidth}{!}{ 
\begin{tabular}{c|l|lllllllllllll}
\toprule
Method &&plane& bcycl& bus &car& horse& knife& mcycl& person& plant &sktbrd &train& truck& Avg     \\
\hline
\hline
ResNet-50&\multirow{8}{*}{\rotatebox{90}{ResNet}}  & 55.1 &53.3& 61.9& 59.1& 80.6 &17.9 &79.7& 31.2& 81.0 &26.5 &73.5& 8.5& \colorbox{lightgray}{52.4} \\
BNM &&89.6 &61.5 &76.9 &55.0& 89.3& 69.1 &81.3 &65.5& 90.0 &47.3& 89.1 &30.1& \colorbox{lightgray}{70.4}\\
MCD& &  87.0 &60.9 &83.7& 64.0& 88.9& 79.6& 84.7& 76.9& 88.6& 40.3& 83.0& 25.8 &\colorbox{lightgray}{71.9}\\
SWD&&90.8 &82.5 &81.7 &70.5& 91.7 &69.5& 86.3& 77.5 &87.4& 63.6& 85.6 &29.2 &\colorbox{lightgray}{76.4}\\
DWL&& 90.7 &80.2& 86.1& 67.6 &92.4 &81.5& 86.8& 78.0& 90.6 &57.1 &85.6& 28.7& \colorbox{lightgray}{77.1}\\
CGDM& & 93.4& 82.7& 73.2& 68.4& 92.9 &94.5 &88.7 &82.1 &93.4 &82.5& 86.8& 49.2& \colorbox{lightgray}{82.3}\\
CAN&&97 &87.2& 82.5 &74.3 &97.8& 96.2 &90.8& 80.7 &96.6& 96.3& 87.5& 59.9& \colorbox{lightgray}{87.2}\\
FixBi && 96.1 &87.8& 90.5& 90.3 &96.8& 95.3 &92.8& 88.7& 97.2 &94.2& 90.9 &25.7& \colorbox{lightgray}{87.2}  \\ 
\midrule
TVT&\multirow{7}{*}{\rotatebox{90}{ViT}}& 82.9 &85.6 &77.5 &60.5& 93.6& 98.2& 89.4& 76.4& 93.6& 92.0& 91.7& 55.7&\colorbox{lightgray}{83.1}\\
Deit-Base & &  98.2 &73.0 &82.5& 62.0 &97.3& 63.5& 96.5& 29.8&68.7 &86.7 &96.7&23.6&\colorbox{lightgray}{73.2} \\
CDTrans-Deit&&97.1& 90.5& 82.4& 77.5& 96.6&96.1 &93.6 &\textbf{88.6}&\textbf{97.9}&86.9 &90.3&\textbf{62.8}&\colorbox{lightgray}{88.4}\\
\textbf{PMTrans-Deit}&&98.2& 92.2 &88.1 &77.0 &97.4&95.8 &94.0&72.1 &97.1 &95.2 &94.6&51.0 &\colorbox{lightgray}{87.7}\\
ViT-Base & &99.1& 60.7& 70.1& 82.7& 96.5 &73.1& 97.1 &19.7 &64.5& 94.7 &97.2& 15.4&\colorbox{lightgray}{72.6}\\
SSRT-ViT&&98.9& 87.6 &\textbf{89.1}&\textbf{84.8}& 98.3& \textbf{98.7}& 96.3&81.1& 94.8 &97.9& 94.5 &43.1& \colorbox{lightgray}{\textbf{88.8}}\\
\textbf{PMTrans-ViT}&& 98.9 & \textbf{93.7} &84.5 &73.3 &\textbf{99.0} &98.0 &96.2 &67.8 &94.2 &\textbf{98.4} &96.6 &49.0 & \colorbox{lightgray}{87.5}\\
\midrule
Swin-Base &\multirow{2}{*}{\rotatebox{90}{Swin}}        &99.3   &  63.4 &  85.9 & 68.9  &  95.1 & 79.6  &\textbf{ 97.1}&29.0&81.4&94.2&\textbf{97.7}&29.6&\colorbox{lightgray}{76.8}  \\
\textbf{PMTrans-Swin}&& \textbf{99.4}                        & 88.3  & 88.1  &78.9   & 98.8  &98.3   &95.8&70.3   &94.6&98.3&96.3&48.5& \colorbox{lightgray}{88.0}  \\ 
\bottomrule
\end{tabular}}
\caption{Comparison with SoTA methods on VisDA-2017. The best performance is marked as \textbf{bold}.}
\label{tab:VisDA_result}
\end{table*}
\begin{table*}[t]
\centering
\resizebox{\linewidth}{!}{ 
\begin{tabular}{c|lllllllllllll}
\hline
backbone &plane& bcycl& bus &car& horse& knife& mcycl& person& plant &sktbrd &train& truck& Avg     \\
\hline
\hline
ViT-bs8 & 99.0 & 92.7 &84.3 &68.0 &\textbf{99.1} &\textbf{98.5} &96.4 &37.6 &93.6 &98.5 & 96.7 & 48.2 & \colorbox{lightgray}{84.4}\\ 
ViT-bs16 & 99.1 & 91.9 &85.9 &69.7 &99.0 &\textbf{98.5} &96.5 &43.1 &93.8 & \textbf{99.2} & \textbf{96.9} &\textbf{50.5} &\colorbox{lightgray}{85.3}\\ 
ViT-bs24 & 98.8 & 92.8 &84.5 &71.1 &\textbf{99.1} &98.3 &\textbf{96.7}&58.9 &93.8 & 98.8 & 96.7 &47.7 &\colorbox{lightgray}{86.4}\\ 
ViT-bs32& 98.9 & 93.7 &84.5 &73.3 &99.0&98.0 &96.2 &67.8 &94.2 &98.4 &96.6 &49.0 &\colorbox{lightgray}{87.5}\\
\hline
Deit-bs8 & 98.1 & 89.5 &86.9 &73.5 &97.5 &96.9 &95.7 &71.8 &96.3 & 92.1 & 95.6 &45.5 &\colorbox{lightgray}{86.6}\\ 
Deit-bs16 & 98.3 & 90.0 &87.0 &74.2 &97.4 &96.9 &95.7 &72.2 &96.7 & 92.2 & 95.8 &46.5 &\colorbox{lightgray}{86.9}\\ 
Deit-bs24 & 98.2 & 90.2 &87.0 &74.8 &97.5 &96.8 &95.7 &\textbf{73.2} &96.8 & 92.1 & 95.6 &46.9&\colorbox{lightgray}{87.1}\\ 
Deit-bs32 &98.2& \textbf{92.2} &\textbf{88.1} &77.0 &97.4 &95.8 &94.0 &72.1 &\textbf{97.1} &95.2 &94.6 &51.0 &\colorbox{lightgray}{87.7}\\

\hline
Swin-bs8 & 99.3 & 87.3 &87.7 &66.9 &98.8 &98.1 &96.4 &57.5 &95.2 & 98.0 & 96.5 &44.2 & \colorbox{lightgray}{85.5}\\ 
Swin-bs16 & 99.2 & 87.6 &87.5 &66.4 &98.8&98.3 &96.3 &58.4 &95.4 & 98.0 & 96.5 &44.6 &\colorbox{lightgray}{85.6}\\ 
Swin-bs24 & 99.2 & 88.1 &87.3 &67.1 &98.7 &98.2 &96.1 &67.1 &94.0 & 97.9 & 96.3 &44.2 &\colorbox{lightgray}{86.2}\\ 
Swin-bs32&\textbf{99.4}& 88.3  & \textbf{88.1}  &\textbf{78.9}   & 98.8  &98.3   &95.8&70.3   &94.6& 98.3&96.3&48.5& \colorbox{lightgray}{\textbf{88.0}}  \\

\hline
\end{tabular}}
\caption{Comparisons between different backbones with different batch sizes on VisDA-2017. The best performance is marked as \textbf{bold}.}
\label{tab:VisDA_batch_size}
\vspace{-10pt}
\end{table*}
\begin{table*}[t]
\centering
\resizebox{\linewidth}{!}{ 
\begin{tabular}{c|llllllllllllll}
\hline
Method&A$\to$ C& A$\to$ P & A $\to$ R & C $\to$ A &C $\to$ P &C $\to$ R &P$\to$ A& P$\to$ C & P$ \to$ R & R$ \to$ A &R$ \to$ C &R$ \to$ P& Avg     \\
\hline
\hline
w/o class information& \textbf{81.3}&\textbf{92.9}&92.8&88.4&93.4&93.2&87.9&\textbf{80.4}&93.0&89.0&\textbf{80.9}&94.8&\colorbox{lightgray}{89.0} \\
w/ class information&80.9&92.7&\textbf{93.4}&\textbf{88.9}&\textbf{93.7}&\textbf{93.9}&\textbf{88.3}&80.2&\textbf{93.5}&\textbf{89.4}&80.3&\textbf{95.3}&\colorbox{lightgray}{\textbf{89.2}}\\
\hline
\end{tabular}}
\caption{Effect of semi-supervised loss with class information. The best performance is marked as \textbf{bold}.}
\label{tab:semi}
\end{table*}
\subsection{Training}
\begin{figure*}
    \centering
    \includegraphics[width=1\linewidth]{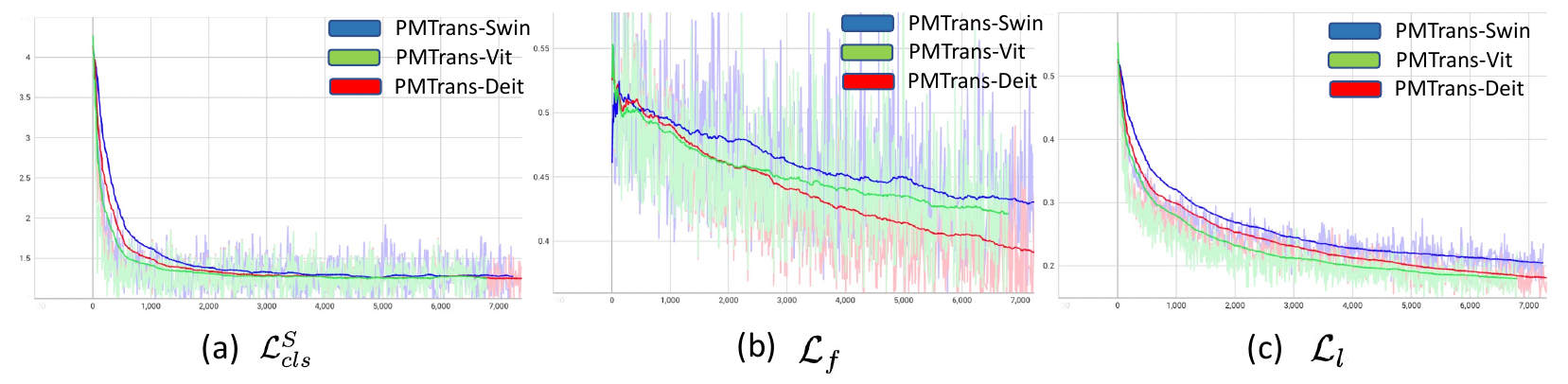}
    \caption{Loss on task A $\to$ C (Office-Home). Lines are smoothed for clarity.}
    \label{fig:loss}
\end{figure*}

We show the progress of training on PMTrans-Swin, PMTrans-Deit, and PMTrans-ViT. To specify how each loss changes, including semi-supervised mixup loss in the label space $\mathcal{L}_l$, semi-supervised mixup loss in the feature space $\mathcal{L}_f$, and source classification loss $\mathcal{L}_{cls}^{S}$, we conduct the experiment on task $A\to C$ on Office-Home for above architectures, and the results are shown in Fig.\ref{fig:loss}. We observe that for all models, both $\mathcal{L}_f$ and $\mathcal{L}_l$ drop constantly, which means the domain gap is reducing as the training evolves. Significantly, $\mathcal{L}_f$ fluctuates more than $\mathcal{L}_l$ as it aligns the domains in the feature space with a higher dimension. 
\subsection{Testing}
In Fig.\ref{fig:my_label}, we testify PMTrans-Swin, PMTrans-ViT, and PMTrans-Deit on the task A $\to$ C on the Office-Home dataset. From Fig. \ref{fig:my_label}, with the same number of epochs, PMTrans-ViT achieves faster convergence than PMTrans-Swin and PMTrans-Deit. Besides, the results further reveal that our proposed PMTrans with different transformer backbones can bridge the source and target domains well and decrease domain divergence effectively.
\subsection{Complexity}
We compare our computational budget with the typical work CDTrans~\cite{abs-2109-06165} on aligning the source and target domains, excluding the choice of backbone. Precisely, CDTrans compute the similarity between patches from two domains by the multi-head self-attention. We are given $n$ as the sequence length, $d$ as the representation dimension, and $c$ as the number of classes. The per-layer complexity is $O(n^2 d)$. While in PMTrans, we adopt CE loss to close the domain gap on both the feature and label spaces of the out, whose complexity is $O(d)+O(c)$. When building the intermediate domain, PatchMix samples patches element-wisely, and its complexity is $O(n)$. As attention scores we use are taken directly from the parameters of Transformer and Classifier, so it brings no additional cost. PMTrans's complexity is $O(d+c+n)$, so it is much more lightweight than the cross attention in CDTrans. 
\subsection{Attention map visualization for target data}
\begin{figure*}[t]
    \centering
    \includegraphics[width=0.8\linewidth]{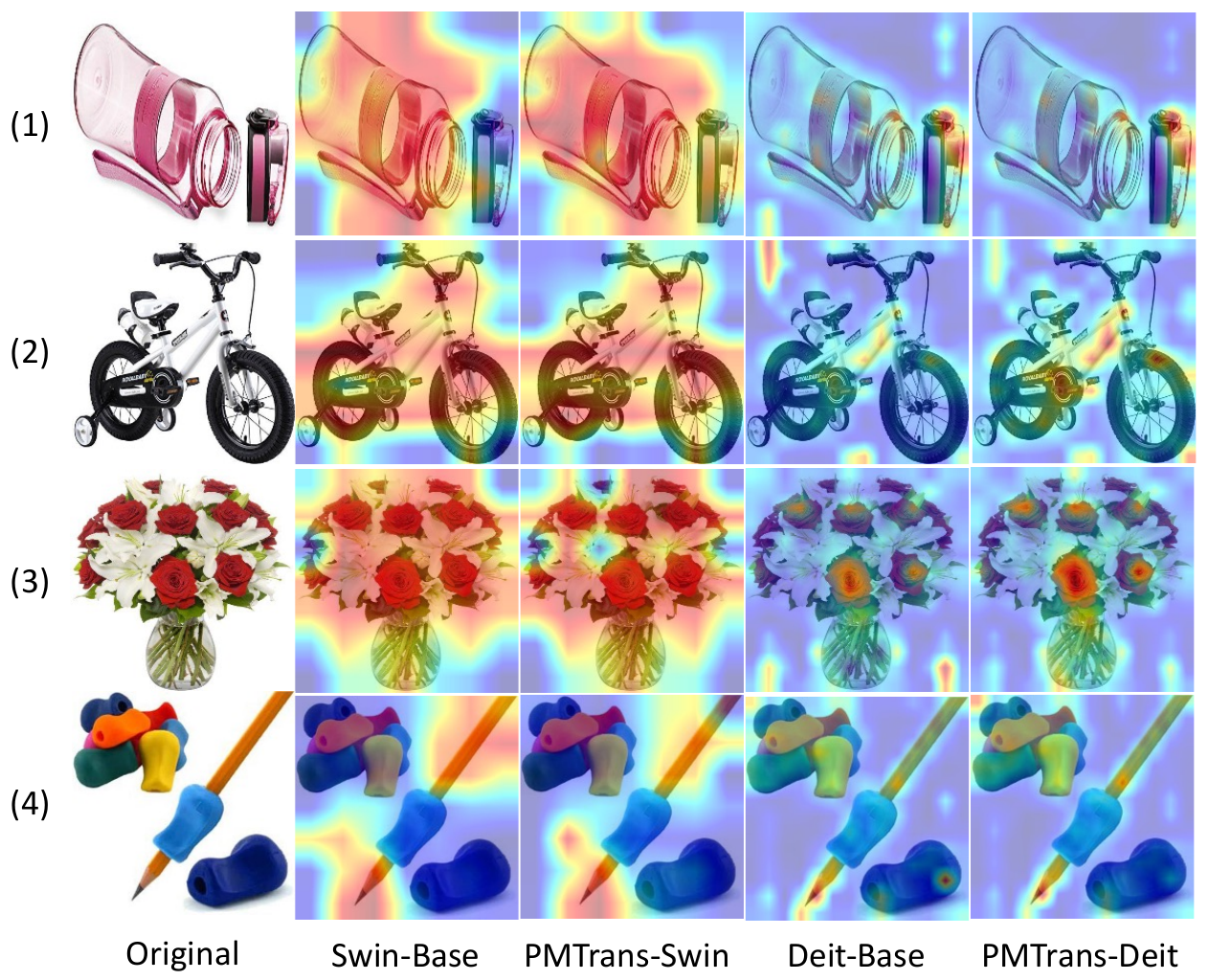}
    \caption{Attention visualization on Swin-based and Deit-based backbones.}
    \label{fig:visual_attention_target}
\end{figure*}

We randomly sample four images from Product (P) of Office-Home and use the pre-trained models $C\to P$ including PMTrans-Swin and PMTrans-Deit to infer the attention maps following the methods described in Sec.\ref{sec:attention map}. In Fig.~\ref{fig:visual_attention_target}, we compare the two PMTrans with their counterparts trained with only source classification loss. We observe that after domain alignment, the attention maps tend to be more focused on the objects \ie less noise around them. Interestingly, for the image whose ground truth label is {\ttfamily pencil} in the fourth row, Swin-based backbone can distinguish it from {\ttfamily plasticine} around or attached to it. At the same time, Deit-based attention covers them all, which may bring negative effects. When the attention scores are used to scale the weights of patches during constructing the intermediate domain in Eq.\ref{eq:lambda}, Swin-based architecture can focus more on semantics while others may not. That may be one of the reasons why PMTrans-Swin gets superior performance on many datasets. Similarly, TS-CAM \cite{DBLP:journals/corr/abs-2103-14862} names the original attention scores from Transformer like Fig.\ref{fig:attention} \textcolor{red}{(a)} as semantic-agnostic, while what we do in Fig.\ref{fig:attention} \textcolor{red}{(b)} is to reallocate the semantics from Classifier back into the patches and make it be aware of specific class activation. 
\section{Ablation Study}
\label{Allation}
\subsection{Batch size}
In Tab. \ref{tab:VisDA_batch_size}, we study the effect of the batch size with different backbones in our proposed PMTrans framework. As shown in Tab. \ref{tab:VisDA_batch_size}, when the batch size is bigger, the input can represent the data distributions better, and therefore the proposed PMTrans based on different backbones with larger batch sizes generally achieves better performance in UDA tasks. Considering the hardware limit, we cannot train models with a batch size of more than 32, so our performance may be lower than it could be, especially when putting in the same condition with a 64 batch size as many previous works do. 

\subsection{Semi-supervised mixup loss with class information}
Tab. ~\ref{tab:semi} shows the comparisons between PMTrans, where the semi-supervised mixup loss combines the class information of target data or not. Note that we use the pseudo labels of target data to calculate the discrepancy between the features and labels. We agree that the semi-supervised mixup loss with class information decreases the domain gaps by reducing the disparity between the feature and label similarities with supervised techniques.

\section{Discussion and Details}
\label{Discussion}
\subsection{Details of mixing labels}
In practice, labels are weighted by attention score, since attention score of high-response regions accounts for most of the context ( as shown in Fig. \textcolor{red}{6}). Therefore, mixing the re-weight labels approximates mixing labels, which is equivalent to mixing patches proved by Theorem.1. Experiments verify that PMTrans with attention score converges easier without sacrificing performance. 
\subsection{Details of semi-supervised losses}
Pseudo-label generation in the target domain is done by every epoch. As the performance of cross-attention in CDTrans highly depends on the quality of pseudo labels, it becomes less effective when the domain gap becomes large. Therefore, we address this issue in three ways: (a) building the intermediate domain; (b) re-weighting intermediate images' labels; (c) generating the pseudo labels of the target at each epoch. This way, the pseudo labels of the target are more reliable as the domain shift decreases, thus enabling the proposed PMTrans to transfer the knowledge between domains well.

\subsection{Details of $y^{it}$ }
Note that $y^{it}$ indeed does not cause any confirmation bias because $y^{it}$ is defined \textit{without considering pseudo labels of target} (as only noisy pseudo labels lead to confirmation bias). \textit{The novelty of our idea lies in that we utilize the identity matrix to measure the label similarity} $y^{it}$ (without using pseudo labels of target) instead of similarity, like $y^{is}$, to decrease the confirmation bias.

\subsection{Learning hyper-parameters of mixup}
\textit{PatchMix mixes patches from two domains for maximizing CE even if $\lambda$ is fixed}. That is, PatchMix can maximize CE, and the other two players minimize CE. More importantly, we propose to \textit{learn the parameters of Beta distribution for the flexible or learnable mixup, so as to build a more effective intermediate domain for bridging the source and target domains}. Numerically, Tab. \textcolor{red}{6} in main paper demonstrates that learning parameters of Beta distribution is \textit{superior} than the fixed parameters in the same min-max CE game.

\subsection{t-SNE visualization}
Source instances are naturally more cohesive than target instances because only source supervision is accessible before adaptation. After adaptation, PMTrans constructs the compact cluster of the target domain instances closer to the source domain instances than Swin-Base. The comparison of visualization proves that PMTrans can effectively bridge the gap between domains.

\subsection{Effectiveness of attention score} 
We conduct an ablation study on Office-Home; and PMTrans-ViT with or without attention score achieves the same performance \textbf{88.9\%}. And results show \textit{PMTrans with attention are easier to converge than that without attention} due PMTrans can utilize attention score to obtain high-response regions.

\subsection{Impart of patch size on performance}
We compare our PMTrans with SSRT using the same batch size as 32 for an apple-to-apple comparison. The result shows that PMTrans achieve comparable performance with SSRT (\textbf{88.0\%} vs. \textbf{88.0\%}), demonstrating the effectiveness of PMTrans.
\clearpage

{\small
\bibliographystyle{ieee_fullname}
\bibliography{egbib}
}